\documentclass[10pt,journal,compsoc]{IEEEtran}
\ifCLASSOPTIONcompsoc
\usepackage[nocompress]{cite}
\else
\usepackage{cite}
\fi
\ifCLASSINFOpdf
\else
\fi
\usepackage{booktabs}
\usepackage{algorithm} 
\usepackage{algpseudocode}
\usepackage{mathrsfs}
\usepackage{multirow}
\usepackage{color}
\usepackage{colortbl}
\usepackage{arydshln}
\usepackage{array}
\usepackage{bm}
\usepackage{rotating}
\usepackage{adjustbox}
\usepackage{color}

\usepackage{textcomp}
\usepackage{mdframed}
\usepackage{amsmath}
\usepackage{CJKutf8}
\usepackage{algorithm} 
\usepackage{balance}
\usepackage{algpseudocode}

\usepackage{tikz}
\newcommand{\coloredcircle}[1][red]{%
    \begin{tikzpicture}[baseline=-0.5ex]
        \fill[#1] (0,0) circle (0.3em);
        \draw[black] (0,0) circle (0.3em);
    \end{tikzpicture}%
}
\usepackage{soul}
\usepackage{amssymb}
\usepackage{multirow}
\definecolor{cvprblue}{rgb}{0.21,0.49,0.74}
\definecolor{myblue}{rgb}{0.97,0.53,0.53}
\definecolor{myred}{rgb}{0.1,0.53,0.99}
\definecolor{hld}{rgb}{0.97,0.81,0.80} 
\definecolor{hlg}{rgb}{1.0,0.90,0.8} 

\definecolor{prompt}{rgb}{0.85,0.91,0.99} 
\definecolor{image}{rgb}{0.8,0.99,0.8} 
\definecolor{condition}{rgb}{1.0,0.8,0.901} 
\definecolor{response}{rgb}{0.88,0.835,0.906} 

\newcommand{\Sm}{\boldsymbol{\mathcal{S}}}

\newcommand{\Cm}{\boldsymbol{C}} 
\newcommand{\hzz}[1]{{\color{black}{#1}}}
\usepackage{amsmath}
\usepackage{amsthm}

\usepackage{multirow}
\usepackage{subfigure}
\usepackage{float}
\usepackage{longtable}
\usepackage{colortbl}
\usepackage{wrapfig}
\usepackage{dsfont}
\usepackage{amssymb}
\usepackage{bbding}

\makeatletter
\let\NAT@parse\undefined
\makeatother
\usepackage{hyperref}

\begin{document}
	%
	\title{A Causality-aware Paradigm for Evaluating Creativity of Multimodal Large Language Models}
	%
	%
	%
	
	\author{Zhongzhan~Huang$^*$,
		Shanshan Zhong$^*$,
		Pan Zhou$^*$,
		Shanghua Gao,
		Marinka Zitnik,
        Liang Lin,\IEEEmembership{Fellow,~IEEE}
		\IEEEcompsocitemizethanks{\IEEEcompsocthanksitem Z. Huang, S. Zhong and L. Lin are with the Sun Yat-sen University, China.  P. Zhou is with the Singapore Management University. S. Gao and M. Zitnik are with the Harvard University, USA. L. Lin is also with Guangdong Key Laboratory of Big Data Analysis and Processing and Peng Cheng Laboratory. $^*$Z. Huang, S. Zhong and P. Zhou have equal contribution for this paper. Corresponding author is L. Lin (e-mail: linliang@ieee.org). 
		}
  }

	\IEEEtitleabstractindextext{%
		\begin{abstract}

Recently, numerous benchmarks have been developed to evaluate the logical reasoning abilities of large language models (LLMs). However, assessing the equally important creative capabilities of LLMs is challenging due to the subjective, diverse, and data-scarce nature of creativity, especially in multimodal scenarios. In this paper, we consider the comprehensive pipeline for evaluating the creativity of multimodal LLMs, with a focus on suitable evaluation platforms and methodologies. First, we find the Oogiri game—a creativity-driven task requiring humor, associative thinking, and the ability to produce unexpected responses to text, images, or both. This game aligns well with the input-output structure of modern multimodal LLMs and benefits from a rich repository of high-quality, human-annotated creative responses, making it an ideal platform for studying LLM creativity. Next, beyond using the Oogiri game for standard evaluations like ranking and selection, we propose LoTbench, an interactive, causality-aware evaluation framework, to further address some intrinsic risks in standard evaluations, such as information leakage and limited interpretability. The proposed LoTbench not only quantifies LLM creativity more effectively but also visualizes the underlying creative thought processes. Our results show that while most LLMs exhibit constrained creativity, the performance gap between LLMs and humans is not insurmountable. Furthermore, we observe a strong correlation between results from the multimodal cognition benchmark MMMU and LoTbench, but only a weak connection with traditional creativity metrics. This suggests that LoTbench better aligns with human cognitive theories, highlighting cognition as a critical foundation in the early stages of creativity and enabling the bridging of diverse concepts. \href{https://lotbench.github.io/}{Project Page.}

		\end{abstract}
		
		}

	\maketitle

	\IEEEdisplaynontitleabstractindextext

	%
	\IEEEpeerreviewmaketitle


\section{Introduction}
\label{sec:intro}

\begin{figure}[t]
  \centering
 \includegraphics[width=0.99\linewidth]{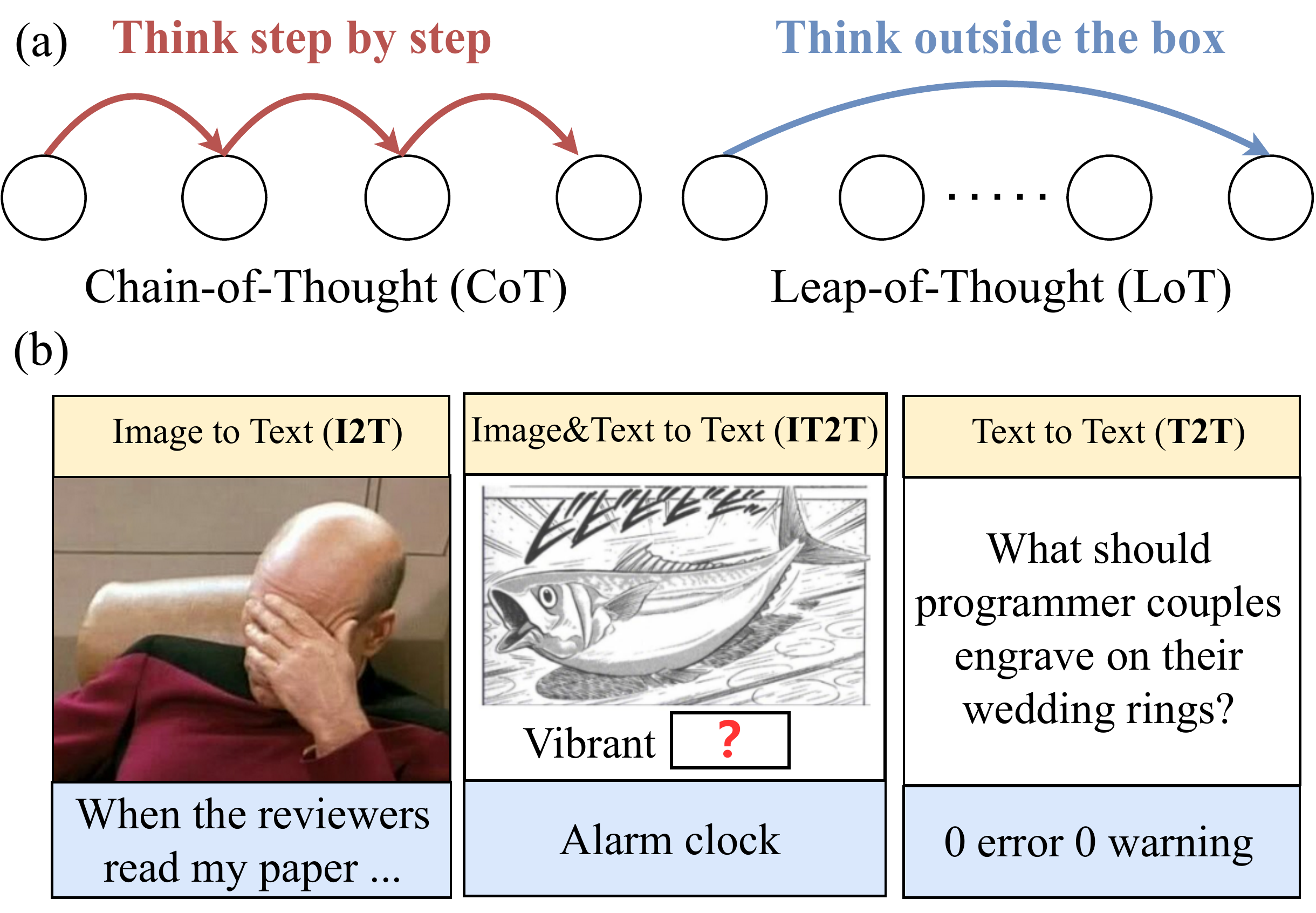}
  \caption{Leap-of-Thought (LoT) for creativity. (a) Comparison of  CoT and LoT.   ``$\bigcirc$" denotes the thought and ``\textrightarrow" represents  the connection between two thoughts. LoT is one of most important ability in creativity~\cite{talmor2020leap,callaway2013cognitive}. (b) Examples of the three types of LoT-based Oogiri games. Players are required to make surprising and creative humorous responses (blue box) to the given multimodal information e.g., images, text, or both. }
\label{fig:oogiri}
\vspace{-10pt}
\end{figure}

\IEEEPARstart{L}{arge} language models (LLMs)~\cite{zhong2024let,bai2023qwen,vicuna2023,wei2021finetuned,saparov2022language,zeng2022socratic,zhong2023adapter} have catalyzed in a transformative era in neural network reasoning, revolutionizing various domains within artificial intelligence. Recently, numerous benchmarks~\cite{guo2023can,liang2024scemqa,lu2022learn,yue2024mmmu,zhao2023survey,li2024survey,chang2024survey} have been proposed to evaluate LLMs' rigorous logical reasoning abilities, spurring the development of methods to enhance these capabilities, particularly the representative Chain-of-Thought (CoT) based methods~\cite{wei2022chain,zhang2022automatic,kojima2022large,yao2023tree,long2023large}. These methods equip LLMs with human-like step-by-step reasoning capacity, enabling them to excel in complex reasoning tasks ranging from language comprehension to visual understanding. As illustrated in Fig.~\ref{fig:oogiri} (a), CoT instills LLMs with a sequential thinking process where each subsequent thought builds upon the previous one. This paradigm enhances precision and rigor in logical processing, making it highly effective for problems requiring closely linked logical reasoning.
While CoT-based methods have proven effective for logical reasoning, they may fall short in capturing another equally important thinking mode: creative reasoning. This limitation stems primarily from their sequential nature. For instance, proving an algebraic inequality often follows a step-by-step CoT process, progressing from one inequality to the next. In contrast, a more creative solution might arise from an intuitive flash, such as a geometric interpretation. This type of insight, known as "Leap-of-Thought" (LoT) or mental leap~\cite{holyoak1995mental,olson1981leap,hofstadter1995review,holyoak1996mental}, represents the art of non-sequential thinking through association, drawing parallels between seemingly unrelated concepts, and facilitating a "leap" in knowledge transfer. Unlike CoT reasoning, LoT, as depicted in Fig.~\ref{fig:oogiri} (a), fosters associative reasoning and encourages thinking outside the box, bridging disparate ideas and facilitating conceptual leaps.
Embracing LLMs with strong LoT abilities can unlock significant potential for creativity, contributing to advancements in creative applications.  However, this crucial aspect of creativity presents unique challenges in benchmarking due to the subjective, diverse, and data-scarce nature of creative responses, especially for the multimodal scenarios \cite{zhong2024let}. This limitation hinders the development of methods to stimulate multimodal LLMs' creative abilities.

 In this paper, we consider the comprehensive pipeline for evaluating multimodal LLMs' (MLLMs) creativity, with a focus on suitable evaluation platform and methodologies. 

(1) \textbf{Suitable evaluation platform}. Thoroughly assessing LoT is challenging due to the complexity of measuring creative thinking~\cite{kitto1994measuring,molle1999eeg,jiang2014development} and the difficulty in gathering pertinent data, since generating novel ideas is challenging, even for humans~\cite{kahneman2011thinking}.
Given these constraints, we propose studying LoT in MLLMs through the lens of Oogiri-style humor generation. Oogiri, a traditional Japanese creative game~\cite{oogiri}, requires participants to provide unexpected and humorous responses to prompts in the form of images, text, or a combination of both, as shown in Fig.~\ref{fig:oogiri} (b). 
This game challenges MLLMs to demonstrate a sudden burst of insight and strong associative thinking, presenting a unique challenge for CoT-based methods. Moreover, the Oogiri game aligns with the input-output paradigm of current MLLMs and, due to its popularity, offers a wealth of high-quality, human-annotated creative responses, making it an ideal platform for exploring LoT ability of MLLMs. Moreover, to investigate the LoT ability of LLMs in the Oogiri game, we initially present the multilingual and multimodal Oogiri-GO dataset 
which comprises more than 130,000 high-quality Oogiri samples in English, Chinese, and Japanese, and curated to prompt textual humor in response to inputs that can be images, text, or both.

(2) \textbf{Suitable evaluation methodologies}. First, following the popular standard LLM benchmark paradigm~\cite{guo2023can,liang2024scemqa,lu2022learn,yue2024mmmu,zhao2023survey,li2024survey,chang2024survey}, we also establish a series of standard LLM evaluations by Oogiri-GO, such as ranking and selection~\cite{lin2021riddlesense,jiang2023brainteaser,zhang2022birdqa,zhong2024let}. We find that even advanced LLMs and reasoning frameworks~\cite{wang2023cogvlm,touvron2023llama,vicuna2023,kojima2022large}, including GPT-4 and CoT, despite their exceptional reasoning capabilities and extensive prior knowledge of various forms of humor~\cite{kojima2022large}, still struggle to demonstrate adequate LoT ability for creative humor generation.
Moreover, while standard evaluations offer simplicity and low assessment costs, we identify inherent risks associated with their use in assessing creativity, such as information leakage and limited interpretability. To address these issues, we first propose training LLMs to assist in generating specific high-quality human-level creative responses (HHCRs). Additionally, we introduce a multi-round interactive~\cite{chen2024weakevalstrongevaluatingelicitinglateral,huang2023lateval} evaluation paradigm, LoTbench. With causal reasoning techniques, LoTbench measures creativity by analyzing the average number of rounds required for an LLM to reach HHCRs. Fewer required rounds indicate higher human-level creativity. LoTbench not only effectively evaluates LLM creativity but also provides interpretable visualizations of the LLM’s innovative thought process during interactions.

The results of LoTbench demonstrate that while most MLLMs exhibit limited creativity, the gap between their creativity and human creativity is not substantial. Current MLLMs show the potential to surpass human creativity.
Furthermore, we observe a strong positive correlation between the results of the well-known multimodal LLM cognition benchmark MMMU~\cite{yue2024mmmu} and LoTbench, but a low correlation with standard creativity evaluation. This indicates that LoTbench’s creativity measurements align more closely with human cognitive theories~\cite{martinsen1994effect,martinsen1993insight,kaufmann1979explorer,runco1995cognition,mednick1962associative}, suggesting that cognition serves as a critical foundation in the early stages of creativity, enabling leaps across diverse conceptual spaces. \hzz{Unlike the conference version \cite{zhong2024let}, we have improved the sampling efficiency of creativity data and proposed a more reasonable causality-aware paradigm for evaluating the creativity of multimodal LLMs.}

In summary, our \textbf{contributions} are as follows:

\noindent \textbf{(i)} We discover the ideal platform for studying the LLMs' creativity, the Oogiri game, and develope a comprehensive standard evaluation pipeline to analyze and discuss how to stimulate LLM creativity.

\noindent \textbf{(ii)} Due the inherent risks in standard creativity evaluations, such as information leakage and limited interpretability, we further propose an interactive, causality-aware benchmark called LoTbench. We find that LoTbench align with human cognitive theories and reveal that while the current LLMs'creativity is not very high, it's close to human levels and has the potential to surpass human creativity.

\section{Related Works}
\label{sec:related}

\noindent\textbf{(1) Multimodal LLMs and their creativity.} Recently, multimodal Language Models~\cite{liu2023improvedllava,chen2023minigptv2,wang2023cogvlm,Qwen-VL} have garnered significant attention, particularly due to their impressive reasoning abilities~\cite{wei2021finetuned,saparov2022language,zeng2022socratic}. Moreover, there is a growing focus on exploring the creativity~\cite{chen2023probing,ling2023unleashing,summers2023brainstorm} of LLMs for applications such as scientific discovery~\cite{park2023papers,liang2021stiffness,huang2023fast}, creative writing~\cite{swanson2021story,chakrabarty2022help,wu2022promptchainer}, etc.

\noindent\textbf{(2) Computational humor} is a branch of computational linguistics and artificial intelligence that uses computers in humor research~\cite{binsted2006computational,xu2024good}, which encompasses various tasks, including humor detection~\cite{shahaf2015inside,tanaka2022learning,xu2022hybrid} and humor generation~\cite{amin2020survey,zhang2020let,hossain2020stimulating}, etc. With the advancement of generative LLMs~\cite{touvron2023llama,wang2023cogvlm,Qwen-VL}, humor generation has become a popular focus while humor generation still faces challenges such as insufficient punchlines~\cite{popova2023does} and limited in multimodal contexts~\cite{chauhan2023mhadig,zhong2024mirror}.

\noindent\textbf{(3) Chain-of-Thought based Methods} \hzz{provide the models with ``chain of thoughts"~\cite{wei2022chain,zhang2022automatic,yao2023tree,long2023large,yuan2024visual,feng2024towards,lyu2023faithful,zheng2023ddcot}, i.e., reasoning exemplars~\cite{wei2022chain}, or a simple prompt ``Let's think step by step"~\cite{kojima2022large}, to encourage LLMs to engage in reasoning rather than simply providing answers directly~\cite{huang2022towards}.}

\section{Evaluation Platform: Oogiri game}
\label{sec:oog}

Unlike most logic reasoning benchmarks~\cite{guo2023can,liang2024scemqa,lu2022learn,yue2024mmmu,zhao2023survey,li2024survey,chang2024survey}, creativity tasks suffer from a severe lack of data and high annotation costs, as it is challenging even for humans to generate a large number of creative responses. Recently, some works~\cite{lin2021riddlesense,jiang2023brainteaser,zhang2022birdqa} have been proposed to study the lateral thinking capabilities of LLMs, but they primarily focus on information in the pure text modality. This makes exploring the creativity of multimodal LLMs a significant challenge, thereby hindering the development of methods to enhance their creativity. Fortunately, in this paper, we find that the Oogiri game serves as an ideal evaluation platform.

 \begin{figure*}[t]
  \centering
 \includegraphics[width=0.99\linewidth]{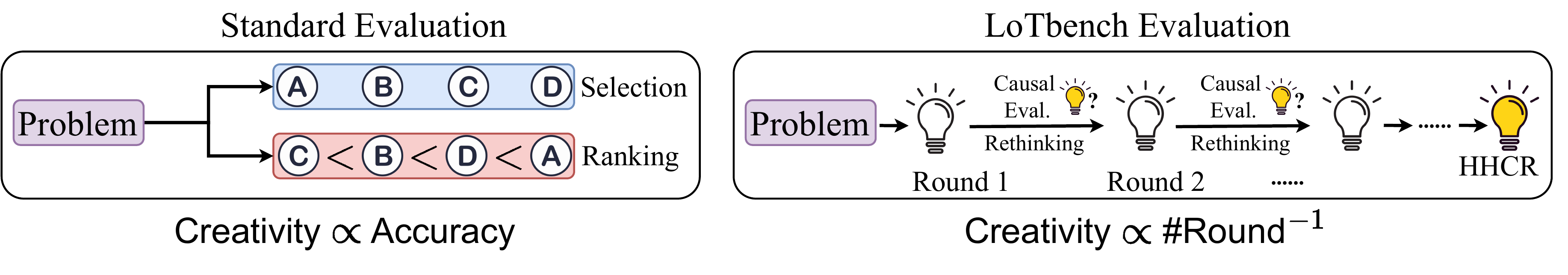}
 \vspace{-15pt}
  \caption{The motivation of different paradigms to measure creativity.
(Left) Standard Evaluation: Assess LLMs by designing selection and ranking tasks. Higher accuracy indicates greater creativity.
(Right) LoTbench: LLMs generate multi-round responses, evaluated by a causal evaluator to determine whether they approach high-quality human-level creative responses (HHCRs). If not, the model enters a rethinking phase for the next round. Creativity is inversely proportional to the number of response rounds (\# Round).}
\label{fig:exameval}
\vspace{-10pt}
\end{figure*}
Oogiri game \begin{CJK*}{UTF8}{goth}(大喜利)\end{CJK*}  is a general term for a series of traditional Japanese comedy games.
In ancient times, there were different types of Oogiri, such as actors performing sumo wrestling, telling ghost stories, etc. The modern Oogiri game mainly refers to one specific type known as  Tonchi \begin{CJK*}{UTF8}{goth}(頓智)\end{CJK*}, typically presented in the format of game shows or intellectual quiz programs~\cite{oogiri}. Players are provided with various multimodal contents, which can be simple questions, random images, etc., and are then prompted to come up with humorous, creative responses to achieve surprising comedic effects, as the examples are shown in Fig.~\ref{fig:oogiri} (b). 
It is worth noting that the character ``\begin{CJK*}{UTF8}{goth}頓\end{CJK*}" in both Japanese and Chinese denote ``sudden", while ``\begin{CJK*}{UTF8}{goth}智\end{CJK*}" means ``intelligence, insight or intuition". This highlights the connection between the Oogiri game and the requirement for strong associative abilities in LoT.
Due to the fact that this creative game aligns with the input-output paradigm of current MLLMs and, because of its popularity, offers a wealth of high-quality, human-annotated creative responses, as well as rich scoring annotations for different responses, such as the number of likes, it makes an ideal platform for exploring the LoT ability of MLLMs.

\subsection{Oogiri-GO Dataset}
\label{sec:oogirigo}

\begin{table}[t]
  \centering
          \caption{Data distribution of the Oogiri-GO dataset.  For the IT2T task, its English version is not available due to cultural preference.}
  \resizebox*{0.85\linewidth}{!}{
    \begin{tabular}{ccccc}
    \toprule
    \textbf{Category} & \textbf{English} & \textbf{Chinese} & \textbf{Japanese}  & \textbf{Total} \\
    \midrule
    I2T   & 17, 336 & 32, 130 & 40, 278  &89, 744 \\
    T2T   & 6, 433  & 15, 797 & 11, 842   &34, 072 \\
    IT2T  & ---     & 912   & 9, 420    &10, 332 \\
    \bottomrule
    \end{tabular}%
    } 
    \vspace{-15pt}
  \label{tab:oogiridata}%
\end{table}%

In this section, we collect Oogiri game data to build a large-scale  Oogiri-GO dataset which serves as the sample of benchmarks to explore the LoT ability.

Specifically, Oogiri-GO is a multimodal and multilingual humor dataset, and contains more than 130,000 Oogiri samples in English, Chinese, and Japanese.  Notably,  in Oogiri-GO, 77.95\% of samples are annotated with human preferences, namely the number of likes, indicating the popularity of a response.  
As illustrated in Fig.~\ref{fig:oogiri} (b),  Oogiri-GO contains three types of Oogiri games according to the input that can be images, text, or both,  and are respectively called   ``Text to Text" (T2T), ``Image to Text"  (I2T),  and ``Image \& Text to Text " (IT2T) for brevity.  Table \ref{tab:oogiridata} summarizes the distribution of these game types.  For training purposes, 95\% of the samples are randomly selected to construct the training dataset, while the remaining 5\% form the test dataset for validation in standard evaluation and analysis.

To create the Oogiri-GO dataset, there are three main steps,  including online data collection, machine filtering by LLM, and manual screening.  Firstly, to collect sufficient data, we source Oogiri game data from the official  Oogiri game platform, Bokete, and other popular platforms, such as  Twitter and Weibo which also host some Oogiri-game-alike data. Then, to guard against the inclusion of bias, violence, explicit content, offensive language, etc., we have placed a strong emphasis on rigorous safety checks during both machine and manual screening.  We first use the multimodal LLM Qwen-VL \cite{Qwen-VL} to do the initial screening of the raw data by constructing safety-checking prompts. Then, manual checking is performed on the remaining data. See more details about the dataset creation in appendix of the conference version~\cite{zhong2024let}.

\section{Standard Evaluation with Oogiri game}
\label{sec:std}

Inspired by the humor benchmarks in \cite{hessel2023androids} and other standard LLM evaluations~\cite{lin2021riddlesense,jiang2023brainteaser,zhang2022birdqa,zhong2024let}, we first develop a standard evaluation, i.e., choice and ranking questions, as shown in Fig.~\ref{fig:exameval} (Left),  and then quantitatively evaluate the LoT ability of LLMs  on the  Oogiri-GO  test dataset.  For the \textit{choice questions}, $m$T$n$ for short, they need LLMs to choose $n$ ``leap-of-thought'' humor responses from $m$  options given the input.  Here we build four types of $m$T$n$ questions, including 2T1, 3T1, 4T1, and 5T2. 2T1 means two options, the ground-truth response (GTR) and an image caption generated by BLIP2 \cite{li2023blip}. 3T1 adds unrelated answers, e.g., other image captions. 4T1 further adds the GTR rewrite by Qwen-14B~\cite{bai2023qwen}. 5T2 has an extra GTR. For these questions, their difficulty increases progressively, and is diverse to ensure comprehensive evaluation. For choice questions, we use accuracy as the evaluation metric. 
Additionally, for the questions in test set whose responses have ground-truth human preference, e.g., the number of likes, we develop the \textit{ranking questions} that always rank five candidates.  For evaluation, we adopt the top-1 accuracy and the widely used ranking metric,i.e., Normalized Discounted Cumulative Gain (NDCG) \cite{jarvelin2002cumulated,radlinski2010comparing}. See more experimental details in the Appendix of conference version~\cite{zhong2024let}.

\begin{figure}[t]
  \centering
 \includegraphics[width=0.9\linewidth]{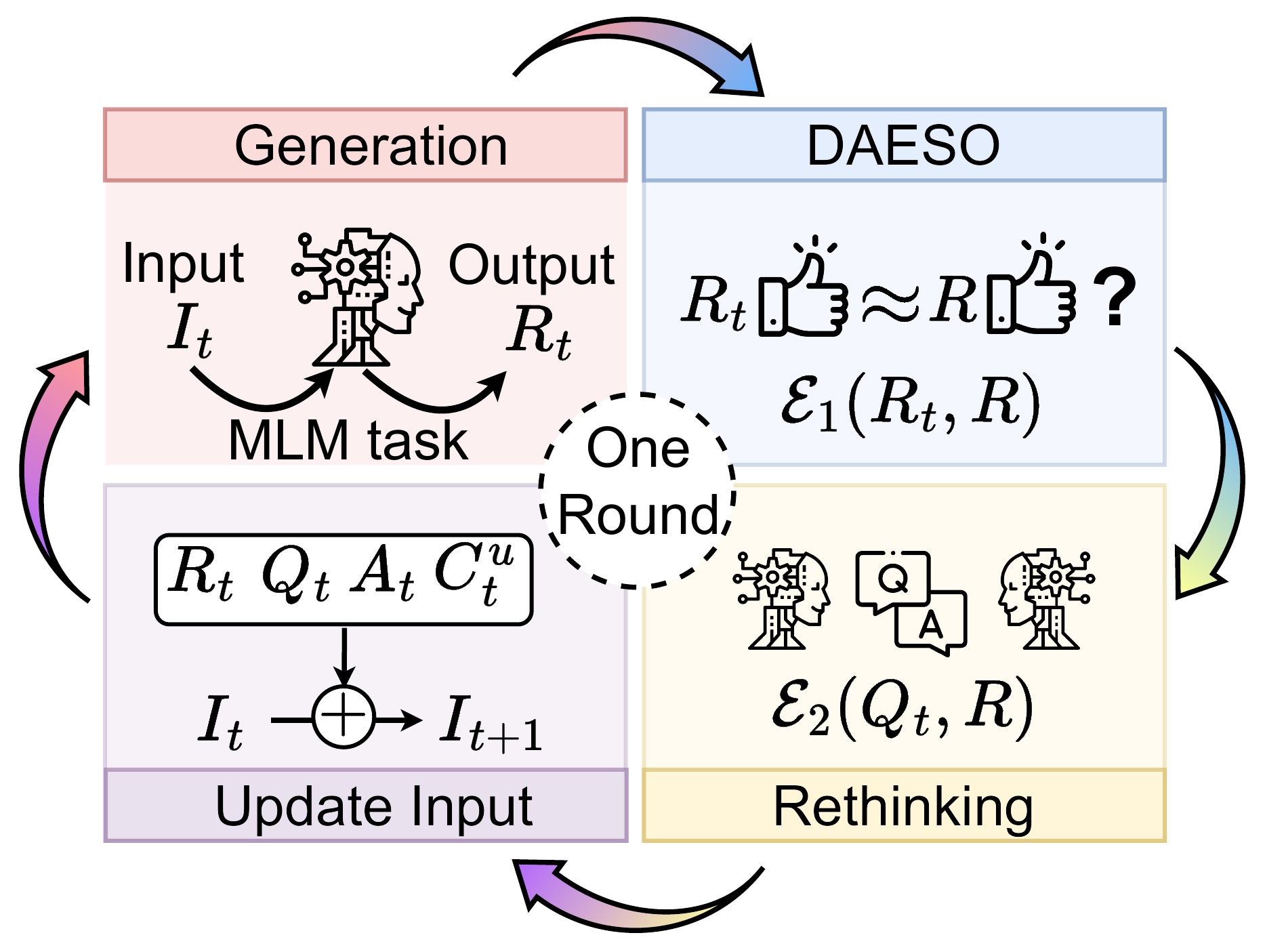}
 \vspace{-10pt}
  \caption{
  	The overview of proposed interactive creativity evaluation LoTbench for LLM. The main task in LoTbench is masked language modeling (MLM) task. DAESO denotes ``different approach but equally satisfactory outcome", where $\mathcal{E}_1$ is the causal evaluator for check whether $R_t$ and $R$ are DAESO. 
  	}
\label{fig:lotbench}
\vspace{-10pt}
\end{figure}

\begin{figure}[t]
  \centering
 \includegraphics[width=0.99\linewidth]{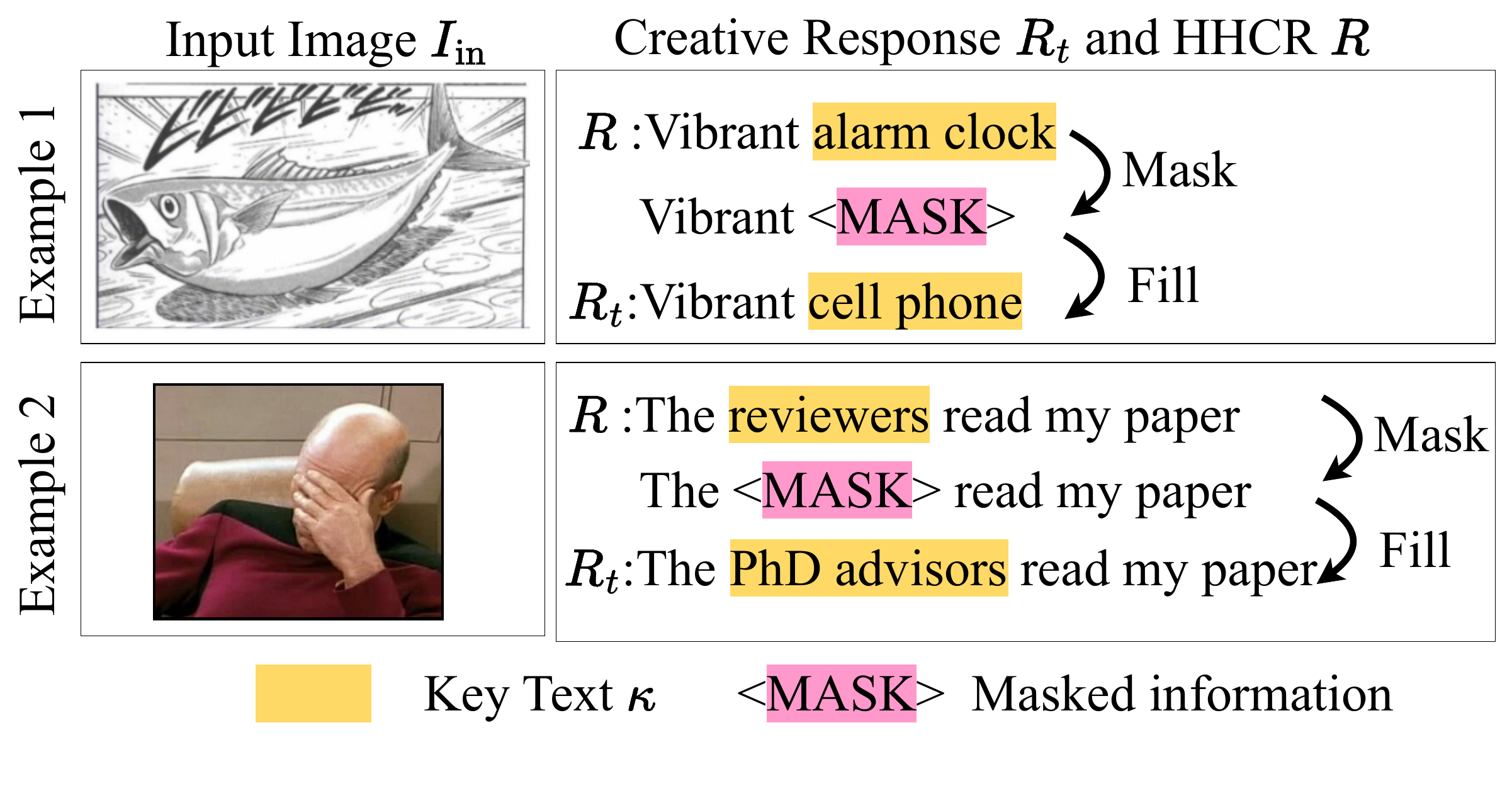}
 \vspace{-10pt}
  \caption{The main task in LoTbench is masked language modeling (MLM).  The LLMs are required to fill in the \textless MASK\textgreater in the sentence to make it a creative response relative to the provided image.}
\label{fig:fill}
\vspace{-10pt}
\end{figure}

\section{LoTbench with Oogiri game}
\label{sec:lotbench_main}

As mentioned in Section \ref{sec:intro}, most existing standard evaluations of LLMs~\cite{lu2022learn,zhao2023survey,chang2024survey,lin2021riddlesense,jiang2023brainteaser,zhang2022birdqa}, including the evaluation presented in Section \ref{sec:std}, are largely based on objective questions such as selection and ranking. These paradigms have significantly contributed to estimating LLM performance and have provided quantitative results, and offer simplicity and low assessment costs. 
However, \textbf{for creativity}, these types of evaluations present certain risks, e.g., limited interpretability and information leakage.

(1) \textbf{Limited interpretability}. Standard evaluation typically adopts the selection and ranking problem format shown in Fig.~\ref{fig:exameval} (Left). Since the LLM directly outputs answers to the questions, we have no insight into the reasoning process behind its ``creative'' decisions. In the experiments in Section \ref{sec:exp}, we also show that many advanced LLMs may exhibit lower performance in standard evaluation scenarios, indicating that there may be some biases when LLMs undergo such tests. However, due to limited interpretability, we cannot trace the reasons for this.

(2) \textbf{Information leakage}.
First, A large amount of information available on the internet, including the Oogiri game, may already have been learned by existing LLMs. 

Moreover, there is the issue of test prompt leakage. For example, in selection evaluations, the correct answer is often easily revealed among the options, which may lead to a less comprehensive assessment of creativity since the evaluation can inadvertently test recognition and logic abilities. For instance, consider the IT2T example in Fig.~\ref{fig:oogiri} (b). For the two options, ``Alarm clock" and "Fish" , the tester can make a judgment \textbf{almost without creativity}, as the former is more unique and interesting. This type of evaluation paradigm does not usually present a problem for non-creativity evaluations, such as for a mathematical reasoning question like "5 + (6 * 4 + 3) = ?", where the options include "32" and "36". The tester must engage in rigorous reasoning to arrive at the correct answer. A truly reasonable creativity evaluation should assess the "measure the creativity level of LLM" rather than "recognize the creativity from LLM."

Facing the issues mentioned above, in this section, we explore the creativity of LLMs from a novel perspective: \textbf{the cost required for LLMs to achieve high-quality human-level creative responses (HHCRs)}. As illustrated in Fig.~\ref{fig:exameval} (Right), we frame this process as an interactive one. Under certain questions, the LLM generates creative responses over multiple rounds and evaluates, using causal inference techniques, whether these responses achieve a different approach but equally satisfactory outcome (DAESO) compared to HHCRs. The fewer rounds required to reach HHCRs, the more creative the LLM is deemed to be, and vice versa.
It is worth noting that since these responses are generated by the LLM itself, ensuring the novelty of the test data can mitigate the problem of information leakage commonly encountered in standard evaluations. Furthermore, this interactive process can effectively visualize the LLM’s innovative thinking process, offering a degree of interpretability.
In Sec. \ref{sec:formulation}, we formalize the LoTbench framework. Next, we introduce the CLoTv2 method in Sec. \ref{sec:syn}, which fine-tunes LLMs to help generate HHCRs in a specific format. In Sec. \ref{sec:testdata}, we detail the construction of HHCRs. Finally, from Sec. \ref{sec:daeso} to Sec. \ref{sec:sc}, we present the other components of LoTbench.

\begin{algorithm}[t]  
    \caption{The details of LoTbench}
    \raggedright 
    \textbf{Input:} Given LLM $\mathcal{A}$ to be tested with a question prompt $Q$ and a generation prompt $G$, along with an independent evaluator $\mathcal{E} = [\mathcal{E}_1, \mathcal{E}_2]$. A  input $I_0 = [I_{\text{in}},\mathcal{C} ]$, where $I_{\text{in}}$ and $\mathcal{C}$ are input image and its caption. A corresponding HHCR $R$. Maximum round $N$. The set of number of round $\mathbf{r}$ and the number of repeated times $m$. The Clue set $C_l = \{C_l^t\}_{t=1}^N$.
    
    \textbf{Output:} Creativity score $S_c$.
        
    \begin{algorithmic}[1]
    
\State{\textbf{While} $m>0$ \textbf{do}}
\State{\indent\textbf{for} $t$ from 0 to $N$ \textbf{do}}
\State\indent\indent Generate response $R_t \gets \mathcal{A}(I_t |G)$ by Sec. \ref{sec:it}
\State\indent\indent Measure $\mathcal{E}_1(R_t, R)$ by Sec. \ref{sec:daeso}
\State\indent\indent \textbf{if} causal evaluator $\mathcal{E}_1(R_t, R)$ is True \textbf{do} break
\State\indent\indent \textbf{if} causal evaluator $\mathcal{E}_1(R_t, R)$ is not True \textbf{do}
\State\indent\indent\indent Ask a question $Q_t \gets \mathcal{A}(I_t,R_t|Q)$ 
\State\indent\indent\indent Get answer $A_t \gets \mathcal{E}_2(Q_t,R)$  by Sec. \ref{sec:qa} 
\State\indent\indent Add $R_t, Q_t, A_t$ and $C_l^t$ into $I_t$ by Sec. \ref{sec:it}
\State {\indent\textbf{end for}}
 
 $m \gets m - 1$ and add $t$ into $\mathbf{r}$
\State {\textbf{end for}}
\State\Return Creativity score $S_c$ with $\mathbf{r}$ by Sec. \ref{sec:sc}
    \end{algorithmic} 
    \label{alg:lotbench}   
\end{algorithm}

 \begin{figure*}[t]
  \centering
 \includegraphics[width=1.00\linewidth]{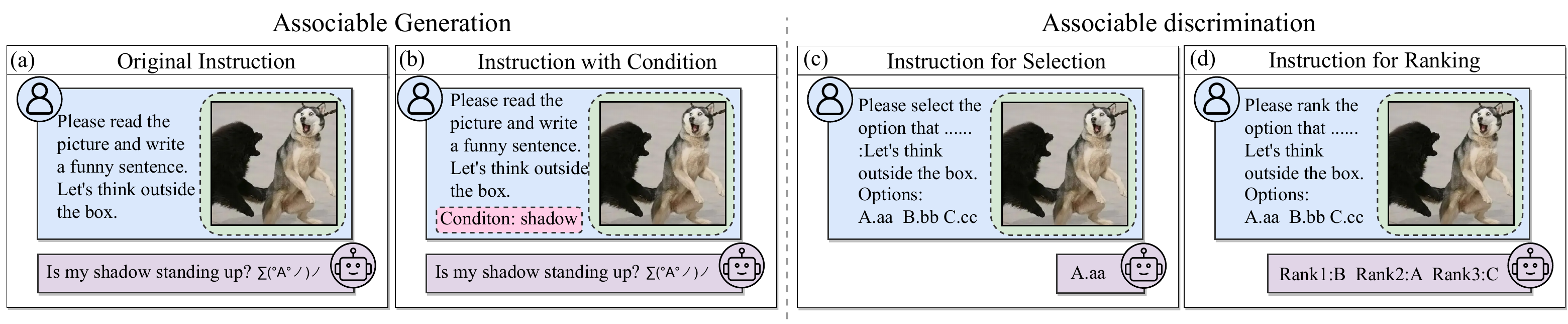}
  \caption{The details of LoT-oriented instructions templates. We take ``Image to Text" as an example, see the Appendix of the conference version~\cite{zhong2024let} for the details of other categories' instructions. (a) and (b) are the instruction templates with/without conditions for associable generation. (c) and (d) are the two instructions about the selection and ranking of associable discrimination. All templates follow the formats in Fig.~\ref{fig:template}.}
\label{fig:format_all}
\end{figure*}
 \begin{figure}[t]
  \centering
 \includegraphics[width=0.9\linewidth]{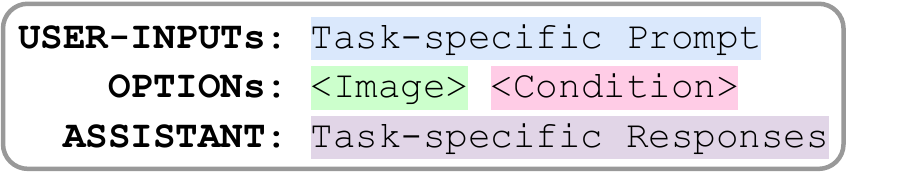}
    \vspace{-0.35cm}
  \caption{The LoT-oriented instruction templates.}
\label{fig:template}
\vspace{-0.3cm}
\end{figure}

\subsection{The formulation of LoTbench}
\label{sec:formulation}

For brevity, we only consider constructing LoTbench with the Chinese and English  data in Oogiri game. Inspired by situation puzzles~\cite{chen2024weakevalstrongevaluatingelicitinglateral,sloane2003leader}, given a input image $I_{\text{in}}$ with its caption $\mathcal{C}$, we ask the LLM to provide an creative response $R_t$ by a masked language modeling (MLM) task as shown in Fig.~\ref{fig:fill} . And in each round of interaction, we determine whether it reaches the creativity level of HHCR $R$ by causal evaluator $\mathcal{E}_1$. Intuitively, fewer rounds imply statistically higher creativity for the LLM. Throughout this process, the LLM can continuously ask questions about $R$ in each round, and the system $\mathcal{E}_2$ will respond with Yes/No. This rethinking of spontaneous questioning  is also a manifestation of its own creativity \cite{sloane2003leader,thinking1970creativity}. To ensure the test can end within a limited number of rounds, we provide the LLM with clues at regular intervals to control its thinking space. The specific algorithm process is shown in Fig.~\ref{fig:lotbench} and Alg.~\ref{alg:lotbench}.

\subsection{Tuning LLM for Data Synthesis by CLoTv2}
\label{sec:syn}

\begin{figure*}[t]
  \centering
 \includegraphics[width=0.99\linewidth]{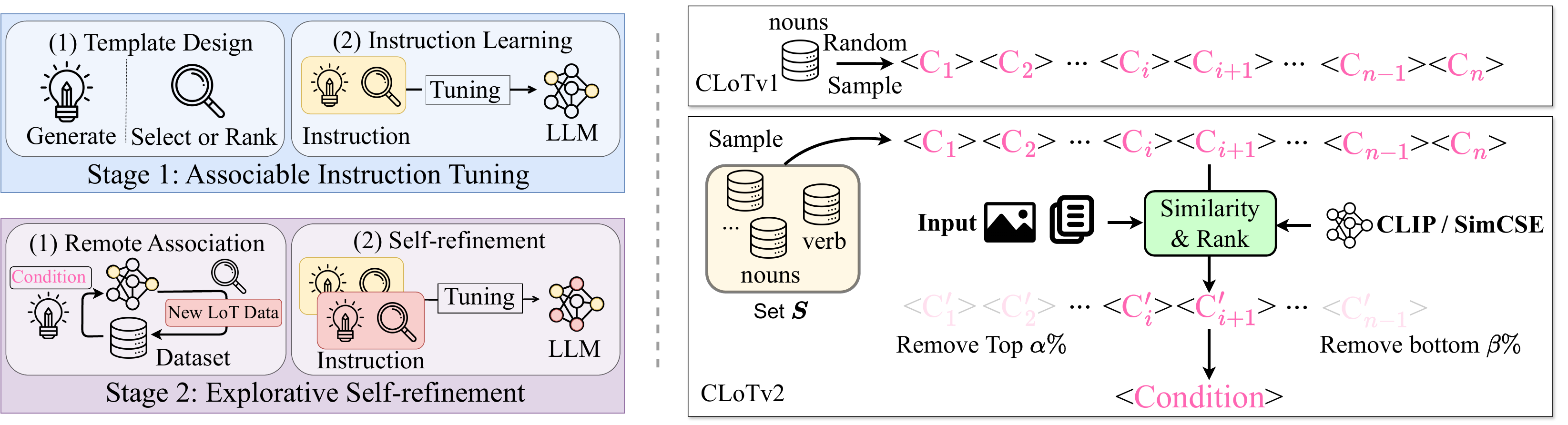}
  \caption{The overview of CLoTv2 to tune LLM for data synthesis. Left: CLoTv2 relies on two  LoT-boosting stages, including associable instruction tuning and explorative self-refinement. Right: the condition sampling method in conference version CLoTv1~\cite{zhong2024let} and CLoTv2.   }
\label{fig:overview}
\end{figure*}

In this section, we tune the LLM through two steps: associable instruction tuning and explorative self-refinement, as shown in Fig.~\ref{fig:overview}, to acquire the ability to generate specific HHCRs in preparation for the test data synthesis of LoTbench in Section \ref{sec:testdata}.
 
\subsubsection{Associable Instruction Tuning}

\label{sec:initial} 
LoT ability mainly includes associable generation and discrimination ability~\cite{Lee2012}. Given an input,  associable generation draws its parallels with seemingly unrelated concepts via remote association and then generates innovative responses, e.g., the unexpected humor for the Oogiri input.  Associable discrimination is to judge the matchiness among input and responses though they are seemingly unrelated, and then to select the most creative response.

Unfortunately, both associable generation and discrimination are not present in current LLMs, e.g., not good performance of  GPT4o~\cite{gpt4} in the Oogiri game observed in Sec.~\ref{sec:exp}. Moreover,  it is hard to improve these two LoT abilities via popular CoT-like prompt techniques. Indeed,  as shown in Sec.~\ref{sec:exp}, CoT even sometimes impairs the LoT performance of the LLMs like Qwen-VL~\cite{Qwen-VL} in the Oogiri game. 
To address this issue, we propose associable instruction tuning which trains LoRA~\cite{hu2021lora} for LLMs on the Oogiri-GO dataset to achieve certain associable generation and discrimination abilities. It has two steps, including instruction generation and discrimination template design, and associable instruction learning.

\noindent \textbf{(1) Instruction Generation \& Discrimination Templates}.  We  design LoT-oriented instruction templates to transform the Oogiri-GO dataset into instruction tuning data, and then train LLM to achieve associable generation and discrimination abilities.    
Our templates primarily comprise two components in Fig.~\ref{fig:template}: task-specific prompt and response.  For different abilities,  the templates need some special design.

For associable generation,   ``USER-INPUTs" contains ``{Task-specific Prompt}" along with two optional   conditions,  ``Image" and ``Condition".  For ``Task-specific Prompt", we elaborately design several templates for different types of Oogiri game. See the Appendix of the conference version~\cite{zhong2024let} for details and there is an image-2-text (I2T) Oogiri example in Fig.~\ref{fig:format_all}.   For ``Image" condition, it relies on the type of Oogiri game, e.g., being the image embeddings in I2T game and empty in T2T type. For the ``condition" option, it's set to empty with a probability of $\rho_c$,  and otherwise is randomly set as one word (including noun, verb, adjective or adverb) in ``task-specific responses". This design gives the  LLM  a clue to connect the game input and the correct responses while also encouraging LLM to explore and unleash its creative thinking with probability $\rho_c$.  Finally,  ``{Task-specific Responses}"  are the ground truth responses of an Oogiri-GO data, and need to be predicted by LLM during training.  This task enforces the LLM to draw parallels between seemingly unrelated concepts in inputs and responses for giving innovative responses, e.g., the humor for the Oogiri input. This associable generation ability can assist the LLM to think outside the box and learn remote association thinking.

Regarding associable discrimination, we aim to develop fundamental LoT discrimination skills for LLM. Based on the Oogiri-GO data, we design choice questions to enhance LLM's LoT discrimination ability, i.e., \textbf{selection} skill. Besides, as 77.95\% of the Oogiri-GO data have human preference annotations, i.e.,  the number of likes of several responses (see Sec.~\ref{sec:oogirigo}),   we design ranking questions to improve another discrimination skill, i,e., \textbf{ranking} ability.

For a choice question, as shown in Fig.~\ref{fig:format_all} (c), the options in ``Task-specific Prompt" contain the random permutations of ground truth response (GTR), image captions generated by BLIP2 \cite{li2023blip},  GTR from other images, rewrites of GTR by Qwen-14B \cite{bai2023qwen}.  See details in Appendix of conference version \cite{zhong2024let}. For ``task-specific responses", it is the GTR. This design is to train LLM to improve its LoT selection ability.  For a ranking question, as shown in   Fig.~\ref{fig:format_all} (d), it is to enforce LLM  to rank multiple distinct responses of a given input to match their human preferences. By training on the choice and ranking questions, LLM is encouraged to distinguish LoT responses and align human creative preferences, improving its   LoT discriminative selection and ranking abilities.

 \noindent \textbf{(2) Associable Instruction Learning}.  By using the above instruction templates,  we augment the  130,000 samples in the Oogiri-GO dataset to more than 600k instructions whose formulation is in Fig.~\ref{fig:template}. During training,  LLM is required to predict the ``task-specific responses" according to the ``USER-INPUTs" which include ``Task-specific Prompt" and two additional optional conditions like image and text condition. To avoid over-fitting, we only train standard LoRA~\cite{hu2021lora}  for the LLM with the associable instruction data. See more details in Appendix of the conference version~\cite{zhong2024let}.

\subsubsection{Explorative Self-Refinement}
\label{sec:remote} 

After associable instruction tuning,  we aim to generate more HHCRs which are then used to train LLM for self-refinement.  To this end, we introduce an innovative stage called explorative self-refinement,   inspired by human LoT exercise process of ``remote association \& self-refinement", also known as mental leap~\cite{Lee2012,holyoak1995mental,holyoak1996mental}.  The remote association process refers to generating new ideas by associating remote concepts or thoughts, and self-refinement uses the generated data to enhance one's own LoT ability. In the following, we design two similar LoT exercise processes for LLM to improve its LoT ability.

 \noindent \textbf{(1) Explorative Remote Association}. The core here is to prompt the LLM to generate a diverse array of creative responses under weakly-associated conditions.

To implement this, as shown in Fig.~\ref{fig:overview} (Right), we first extract a set of keywords, including nouns, verbs, adjectives, and adverbs, denoted as $\Sm$, from the text in the Oogiri-GO training data.
Then, for given image and text in each sample, we construct a series of effective weakly-associated conditions as follows. We sample $n$ candidate conditions $\{\Cm_i\}_{i=1}^n$ from $\Sm$ with equal probability, then use CLIP or SimCSE \cite{radford2021learning,gao2021simcse} to batch compute the similarity between these candidates and the image or text in samples. The similarities of I2T and IT2T are determined by their CLIP scores based on their images and keywords, while SimCSE is used to calculate the similarity for T2T. The candidates are ranked by similarity in descending order to get $\{\Cm_i^\prime\}_{i=1}^n$. We remove the top $\alpha\%$ and bottom $\beta\%$ of elements from $\{\Cm_i^\prime\}_{i=1}^n$, and add an empty condition $\phi$ to obtain the final weakly-associated conditions:
\begin{equation}
     \Cm_{W} = \{\Cm_i^\prime\}_{i=\lfloor \alpha\% n \rfloor}^{\lfloor (1-\beta\%) n \rfloor + 1} \cup \{\phi\}.
 \end{equation}
Next, we add each condition from $\Cm_{W}$ into the instruction shown in Fig.~\ref{fig:template} and feed them into the LLM to generate a humor candidate. We mix each generated humor candidate with its corresponding ground truth responses (GTR), and select the top-1 as the final response using the selection ability learned in Sec.~\ref{sec:initial}. Finally, if the selected top-1 response is the GTR, we discard this generated humor candidate. By repeating this process, we progressively gather sufficient new HHCRs.

 The core of this approach is the selection of weakly-associated conditions, which can encourage the LLM to engage in remote associations. This is because the empty conditions allow LLM to operate freely, while the other conditions compel the LLM to draw connections between seemingly unrelated concepts. This mechanism facilitates the establishment of links between seemingly-unrelated and weakly-related concepts, encouraging the LLM to explore knowledge outside of traditional cognitive limitations. The exploration ability distinguishes our CLoTv2 from CoT which primarily guides the LLM to exploit its inherent reasoning ability without emphasizing knowledge exploration.

Unlike the conference version CLoTv1~\cite{zhong2024let}, which approximates weakly-associated conditions by relying solely on random sampling, as shown in Fig.~\ref{fig:overview} (Right), CLoTv2 explicitly models the conditions sampling based on the similarity between the content of each sample and the candidate conditions. By using parameters $\alpha$ and $\beta$, it ensures that the conditions are sufficiently different from the content but not completely unrelated—achieving a truly "weakly-associated" state. This approach mitigates the issue in CLoTv1, where random sampling often leads to the generation of many irrelevant conditions, which typically fail to produce effective responses, resulting in considerable computational waste. In this paper, we set $n=100$, $\alpha = 25$ and $\beta = 70$.

  \noindent \textbf{(2) Self-refinement}. We combine the above generated instructions with vanilla instruction tuning samples in Sec.~\ref{sec:initial} to form a dataset with more than 660k samples to train our LLM again. Since the above generated data is of high diversity because of its exploration strategy,  they prevent performance collapse~\cite{hataya2023contamination,shumailov2023model} during this phase.

After the two LoT-boosting phases above, the LLM gains sufficient LoT  ability and can assist us in synthesizing new HHCRs to construct the test data for LoTbench in Section \ref{sec:testdata}, which can mitigate the issue of imformation leakage.

\subsection{The Data Construction in LoTbench}
\label{sec:testdata}

\noindent\underline{\textbf{Task type}}. 
\hzz{The primary task in LoTbench is a masked language modeling (MLM) task, as illustrated in Fig.\ref{fig:fill}. Unlike I2T tasks that directly generate a complete response, MLM is a variation of IT2T. This design choice is guided by two key considerations: 
(1) To ensure the task leverages LLM's core strengths. Otherwise, limitations in some specific capabilities might lead to mediocre performance, interfering with the assessment of LLM creativity. MLM is precisely the type of task where LLMs excel~\cite{micheletti2024exploration}; (2) To simplify evaluation complexity. Since creativity is inherently diverse, allowing LLMs to freely generate responses $R_t$ as in I2T tasks would make it difficult to assess whether they match the creativity level of given HHCR $R$ due to high variability. Therefore, some constraints on $R_t$ are necessary, and MLM tasks naturally provide this by fixing certain textual content, making it a suitable choice.

Specifically, as shown in Fig.~\ref{fig:fill}, we manually annotate the key text $\kappa$ in each HHCR $R$, mask it, and ask the LLM to complete the response, aiming for creative and high-quality responses. The "key text" $\kappa$ refers to some textual contents that most crucially link the image and response, making the responses creative. Removing these contents would strip the text-image combination of its creativity. For example, in Fig.~\ref{fig:fill} Example 1, "alarm clock" is the key text in $R$. Moreover, identifying such key text accurately is challenging for different automated tools, including LLMs, so in this paper, we manually annotate them one by one during the data construction process.}

\begin{figure*}[t]
  \centering
 \includegraphics[width=0.99\linewidth]{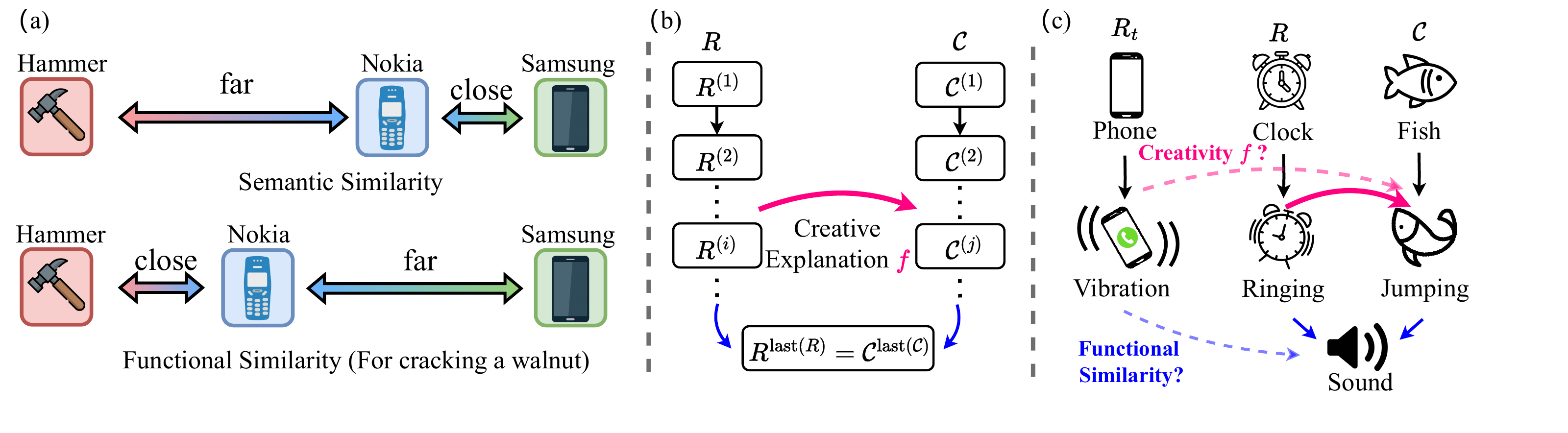}
 \vspace{-10pt}
  \caption{The overview of DAESO. (a) The difference between semantic similarity and functional similarity. (b) The mathematical modeling for DAESO by causal chains of given HHCR $R$ and corresponding image caption $\mathcal{C}$. (c) Two criteria of DAESO. }
\label{fig:nokia}
\vspace{-10pt}
\end{figure*}

\noindent\underline{\textbf{Data structure}}. Each sample in LoTbench consists of six parts: the input image $I_{\text{in}}$ with its corresponding HHCR \( R \), image caption \( \mathcal{C} \), and key text \( \kappa \). It also includes a detailed explanation \( E_{xp} \) of why each \( R \) is innovative, and a Clue set \( C_l = \{C_l^t\}_{t=1}^N \) designed to help LoTbench complete the evaluation within a limited number of rounds. For instance, in the sample shown in Fig.~\ref{fig:fill} Example 1, the input image $I_{\text{in}}$ is the one on the left, with \( R \) being "Vibrant alarm clock", \( \mathcal{C} \) being "A freshly caught fish, still flopping on the table, made a loud noise", and \( \kappa \) being "alarm clock". The explanation \( E_{xp} \) is "The lively fish rapidly flopping on the table and making a lot of noise closely resembles the moment when an alarm clock goes off. In \( R \), the visual association of imagining the fish's flopping as an alarm clock ringing is both surprising and intriguingly interesting." Additionally, we have structured the Clue set \( C_l = \{C_l^t\}_{t=1}^N \) to include both substantive clues and empty clues. At regular intervals, such as every several rounds (set as 5 in our paper) as indicated in line eight of Alg.~\ref{alg:lotbench}, a substantive clue is added to the user-input. An example of a substantive clue is "It is a noun; It is a commonly used object at home; Pay attention to rapid jumping; Related to sound; Related to time."

\noindent\underline{\textbf{Data volume}}. 

Due to some unavoidable reasons, the data volume of LoTbench’s test samples is limited: (1) There is a shortage of data suitable for constructing MLM tasks. On one hand, as mentioned in Section \ref{sec:intro}, creativity data itself is rare, and HHCR data is even scarcer due to the high-quality requirements. Additionally, even when some HHCR data is available, responses in the Oogiri game are typically very short, with entire or major portions often consisting of key text, making it challenging to create MLM tasks. (2) Fairness of the benchmark also needs consideration. Given the nature of the Oogiri game, many HHCR key texts are often culturally or knowledge-specific, so we must filter out these examples to ensure fair evaluation across most LLMs.

For these intrinsic limitations above in creativity data, we carefully and manually curated 106 HHCR samples suited for LoTbench, with Oogiri-GO and the help of CLoTv2 trained in Section \ref{sec:syn} to generate brand-new HHCRs that meet MLM requirements. While LoTbench contains fewer samples than typical LLM evaluations~\cite{guo2023can,liang2024scemqa,lu2022learn,yue2024mmmu,zhao2023survey,li2024survey,chang2024survey}, it boasts extremely high quality. Prior work \cite{polo2024tinybenchmarks,kipnis2024texttt} has emphasized that test data quality is far more important than quantity, noting that the sample in many well-known benchmarks contain severe redundancy
 and a small number of test cases—less than 1\%—can also yield evaluation results comparable to a full dataset. The consistency with human cognitive theories~\cite{martinsen1994effect,martinsen1993insight,kaufmann1979explorer,runco1995cognition,mednick1962associative} demonstrated in LoTbench's evaluation results, shown in Section \ref{sec:result}, also indicates that the current data construction is sufficient for assessing LLM creativity to a certain extent.

\subsection{The Details of $I_t$ and $R_t$}
\label{sec:it}

In Alg.~\ref{alg:lotbench}, $I_t$ initially includes only the input image $I_{in}$ and its corresponding image caption $\mathcal{C}$. The generation prompt $G$ contains the complete instruction, including the system prompt, example prompt for in-context learning, and task-specific prompts. Additionally, $G$ also includes $R$ with the key text $\kappa$ masked. In each round of evaluation, the LLM under test is required to creatively fill in the masked key text in $R$ through the given $I_t$ and prompt $G$. As the LoTbench interactive evaluation progresses, $I_t$ will continuously incorporate the current round's generated response $R_t$, the obtained clue $C_l^t$, and the question and answer $Q_t$ and $A_t$. In the next round, this historical information will help the LLM to further produce a creative response. See supplementary for all prompts and other details.

\subsection{To Measure DAESO with Causal Evaluator $\mathcal{E}_1$?}
\label{sec:daeso}

\subsubsection{Criteria}

In Alg.~\ref{alg:lotbench}, we need an evaluator $\mathcal{E}_1$ to determine whether $R_t$ and a given HHCR $R$ exhibit a similar level of creativity. On one hand, since creativity is diverse, $R_t$ and $R$ are unlikely to be identical at the character level, so $\mathcal{E}_1$ cannot assess them through string matching. On the other hand, we also cannot rely solely on semantic similarity, as is common in natural language processing \cite{gao2021simcse}. For example, as illustrated in Fig.~\ref{fig:fill} Example 1, if $R$ is "vibrant alarm clock" and an LLM outputs $R_t$ as "vibrant cell phone," we may still consider $R_t$ to have a similar level of creativity to $R$ despite the semantic distance between "cell phone" and "alarm clock". 
Through the analysis above, in this paper, the $\mathcal{E}_1(R_t, R)$ is set to assess whether $R_t$ and $R$ are a "different approach but equally satisfactory outcome" (DAESO). To achieve this, we propose two criteria for DAESO: (1) $R_t$ and $R$ share the same creative point; (2) $R_t$ and $R$ are functionally similar rather than semantically similar.

For criterion (1), if $R_t$ is "vibrant drum," even though a drum can also make sound, it lacks the vivid, jump-out image of an "alarm clock" in the given context, thus differing in innovation point. For criterion (2), The DAESO between "cell phone" and "alarm clock"  should be judged functionally rather than semantically. As shown in Fig.~\ref{fig:nokia} (a), if we compare "Nokia", "Samsung" and "Hammer", "Nokia" and "Samsung" would seem closer semantically, as both are well-known electronics companies with popular phone products. However, in a given scenario like "cracking a walnut", "Nokia" and "Hammer" are more similar functionally, as "Nokia" products are famously durable and can serve to crack walnuts, while "Samsung" devices are more fragile and thus less suitable for the task.

Based on these criteria, we propose a novel evaluation mechanism for $\mathcal{E}_1$ in following Section \ref{sec:modeling}, which assesses DAESO by analyzing the causal chain in the LLM responses to judge whether $R_t$ and $R$ align in terms of DAESO.

\subsubsection{Modeling}

\label{sec:modeling}
\noindent\underline{\textbf{Causal Construction}}. For a given sample, we can first leverage the image caption $\mathcal{C}$ and carefully annotated explanation \( E_{xp} \) of "Why \( R \) is creative," as mentioned in Section \ref{sec:testdata}, to model the causal chain for \( R \) and \( \mathcal{C} \). \( R \) and \( \mathcal{C} \) are expanded in the form of Eq.~(\ref{eq:zk}) as follows:
\begin{equation}
\begin{aligned}
& \mathcal{C} \simeq\left(\mathcal{C}^{(1)}, \mathcal{C}^{(2)} \cdots, \mathcal{C}^{\left(\text{last}(\mathcal{C})\right)}\right) \\
& R \simeq\left(R^{(1)}, R^{(2)}, \cdots, R^{\left(\text{last}(R)\right)}\right)
\end{aligned},
\label{eq:zk}
\end{equation}
where \( \mathcal{C}^{(i)} \) and \( R^{(j)} \) represent individual nodes within \( \mathcal{C} \) and \( R \), respectively, with Fig.~\ref{fig:nokia} (b) illustrating a diagram. \( \text{last}(\mathcal{C}) \) and \( \text{last}(R) \) denote the number of nodes in chain of \( \mathcal{C} \) and \( R \), respectively, and they may not be equal. First, for criterion (1), there should be a creative explanation \( f \) and \( i \leq \text{last}(R) \), \( j \leq \text{last}(\mathcal{C}) \), ensuring that:
\begin{equation}
    f(R^{(i)}) \to \mathcal{C}^{(j)},
    \label{eq:rc}
\end{equation}
i.e., there exists a function \( f \) such that one node in \( R \) can be mapped to another node in \( R \), and this mapping \( f \) is the reason why \( R \) is considered creative. Fig.~\ref{fig:nokia} (c) provides a specific analysis for Fig.~\ref{fig:fill} Example1 where the "alarm clock" appears creative and visually engaging because its bouncing motion when ringing corresponds to the bouncing of a "fish", thus creating a vivid and creative image.
Furthermore, for criterion (2), we can consider the final nodes in $\mathcal{C}$ and $R$ to be the same, that is,
\begin{equation}
    \mathcal{C}^{\text{last}(\mathcal{C})} = R^{\text{last}(R)},
\end{equation}
to represent that the functions of the two causal chains are similar. For example, in the case shown in Fig.~\ref{fig:nokia} (c), whether it is  "clock ringing" or "fish jumping", they all functionally serve to "produce sound."

\begin{table*}[htbp]
		\caption{The accuracy (\%) of choice questions and the NDCG (\%) of ranking questions on  \textbf{mutilmodal multilingual models}. $m$T$n$ choice question selects $n$ correct answers from  $m$  options. ``Avg.'' is the average of all metrics. ``AIT" and ``AITv2" denotes the the LLM with only associable instruction tuning of CLoTv1 and CLoTv2, respectively. The best results for each backbone are highlighted in bold and  the second-best results are emphasized with some underlines.
 \vspace{-5pt}
}
	\resizebox*{0.95\linewidth}{!}{
    \begin{tabular}{lr|ccccc|ccccc|ccccc}
    \toprule
    \multirow{2}[4]{*}{Model} & \multirow{2}[4]{*}{Size} & \multicolumn{5}{c|}{Image\&Text to Text (IT2T)} & \multicolumn{5}{c|}{Image to Text (I2T)}    & \multicolumn{5}{c}{Text to Text (T2T)} \\
\cmidrule{3-17}          &       & 3T1   & 4T1   & 5T2   & Rank  & Avg.$\qquad$  & 3T1   & 4T1   & 5T2   & Rank  & Avg.$\qquad$  & 3T1   & 4T1   & 5T2   & Rank  & Avg.$\qquad$ \\
    \midrule
    GPT4v~\cite{gpt4} & -     & 19.3  & 14.9  & 3.2   & 56.7  & 23.5$\qquad$  & 29.1  & 15.1  & 3.9   & 60.4  & 27.1$\qquad$  & 27.1  & 16.8  & 6.8   & 53.5  & 26.1$\qquad$  \\
    LLaVA-1.5 \cite{liu2023improvedllava} & 13B   & 13.2  & 13.7  & 13.9  & 68.1  & 27.2$\qquad$  & 29.3  & 22.7  & 3.9   & 60.9  & 29.2$\qquad$  & 33.8  & 25.2  & 4.0   & 62.6  & 31.4$\qquad$  \\
    MiniGPT-v2~\cite{chen2023minigptv2} & 7B    & 6.1   & 3.4   & 4.0   & 60.7  & 18.6$\qquad$  & 5.3   & 4.0   & 3.8   & 60.5  & 18.4$\qquad$  & 10.8  & 7.3   & 3.5   & 59.4  & 20.3$\qquad$  \\
    mPLUG-Owl$_{\text{Multilingual}}$~\cite{ye2023mplug} & 7B    & 28.1  & 26.0  & 10.5  & 64.4  & 32.2$\qquad$  & 19.2  & 18.6  & 6.0   & 60.5  & 26.1$\qquad$  & 24.4  & 22.2  & 10.7  & 60.1  & 29.4$\qquad$  \\
    VisualGLM-6B~\cite{du2022glm} & 6B    & 24.1  & 22.5  & 9.7   & 67.4  & 30.9$\qquad$  & 14.3  & 20.4  & 8.8   & 61.9  & 26.4$\qquad$  & 13.1  & 20.2  & 7.1   & 61.3  & 25.4$\qquad$  \\
    GPT-4o~\cite{gpt4} & -     & 26.5  & 20.1  & 8.9   & 60.7  & 29.1$\qquad$  & 22.6  & 18.6  & 11.9  & 60.4  & 28.4$\qquad$  & 30.6  & 24.2  & 11.5  & 59.4  & 31.4$\qquad$  \\
    GPT-4o mini~\cite{gpt4} & -     & 19.8  & 24.6  & 14.4  & 67.4  & 31.6$\qquad$  & 25.6  & 22.1  & 8.8   & 61.2  & 29.4$\qquad$  & 32.6  & 25.8  & 13.9  & 61.8  & 33.5$\qquad$  \\
    Claude 3.5 Sonnet & -     & 20.8  & 15.9  & 9.7   & 64.4  & 27.7$\qquad$  & 19.2  & 18.6  & 10.6  & 60.5  & 27.2$\qquad$  & 20.6  & 25.2  & 8.6   & 60.6  & 28.8$\qquad$  \\
    Gemini 1.5 Pro~\cite{gemini} & -     & 18.6  & 20.4  & 10.6  & 66.1  & 28.9$\qquad$  & 20.5  & 16.6  & 6.6   & 60.4  & 26.0$\qquad$  & 26.4  & 15.6  & 5.6   & 62.6  & 27.6$\qquad$  \\
    Intern-VL2~\cite{chen2024internvl} & 40B   & 8.2   & 11.6  & 3.2   & 60.7  & 20.9$\qquad$  & 14.3  & 8.8   & 3.8   & 61.4  & 22.1$\qquad$  & 20.8  & 16.8  & 7.1   & 59.4  & 26.0$\qquad$  \\
    miniCPM-V~\cite{yao2024minicpm} & 8B    & 20.6  & 18.6  & 10.6  & 64.4  & 28.6$\qquad$  & 19.2  & 18.6  & 8.6   & 59.8  & 26.6$\qquad$  & 30.6  & 28.6  & 10.7  & 61.3  & 32.8$\qquad$  \\
    Yi-VL~\cite{young2024yi} & 34B   & 19.3  & 16.6  & 6.8   & 59.6  & 25.6$\qquad$  & 20.4  & 14.8  & 6.6   & 59.4  & 25.3$\qquad$  & 16.5  & 10.5  & 6.8   & 55.6  & 22.4$\qquad$  \\
    Qwen2-VL~\cite{wang2024qwen2} & 72B   & 28.6  & 20.4  & 8.8   & 66.6  & 31.1$\qquad$  & 24.6  & 20.5  & 10.2  & 60.6  & 29.0$\qquad$  & 16.8  & 25.2  & 6.8   & 61.3  & 27.5$\qquad$  \\
    \midrule
    Qwen-VL~\cite{Qwen-VL} & 7B    & 30.2  & 26.0  & 10.4  & 67.7  & 33.6$\qquad$  & 23.2  & 23.1  & 11.9  & 62.2  & 30.1$\qquad$  & 23.4  & 25.0  & 13.3  & 59.6  & 30.3$\qquad$  \\
    Qwen-VL$_{+\text{AITv1}}$ \cite{zhong2024let} & 7B    & 39.7  & \ul{38.9} & 15.7  & 67.3  & 40.4{$_{+\ 6.8}$}  & 38.8  & 30.5  & 15.7  & 62.3  & 36.8{$_{+\ 6.7}$}  & 30.6  & 28.7  & 16.7  & 62.6  & 34.6{$_{+\ 4.3}$}  \\
    Qwen-VL$_{+\text{CLoTv1}}$ \cite{zhong2024let} & 7B    & \ul{41.8} & 38.7  & \ul{21.6} & \ul{68.5} & \ul{42.7}{\color{black}$_{+\ 9.1}$} & \ul{39.8} & \ul{35.1} & \ul{22.7} & \ul{64.4} & \ul{40.5}{\color{black}$_{+10.4}$} & \ul{38.8} & \ul{29.4} & \ul{21.0} & \ul{64.7} & \ul{38.5}{\color{black}$_{+\ 8.2}$} \\
     Qwen-VL$_{+\text{AITv2 }}$ (ours) & 7B    & 40.1  & 39.1 & 17.0  & 67.9  & 41.0{$_{+\ 7.4}$}  & 39.2  & 32.5  & 17.2  & 61.8  & 37.7{$_{+\ 7.6}$}  & 32.6  & 28.5  & 18.3  & 63.1  & 35.6{$_{+\ 5.3}$}  \\
     Qwen-VL$_{+\text{CLoTv2 }}$ (ours) & 7B    & \textbf{42.2} & \textbf{39.2}  & \textbf{22.3} & \textbf{69.0} & \textbf{43.2}{\color{black}$_{+\ 9.6}$} & \textbf{40.4} & \textbf{37.1} & \textbf{24.6} & \textbf{65.2} & \textbf{41.8}{\color{black}$_{+11.7}$} & \textbf{39.2} & \textbf{29.8} & \textbf{23.1} & \textbf{65.3} & \textbf{39.4}{\color{black}$_{+\ 9.1}$} \\
    \bottomrule
    \end{tabular}%
	}
 \centering

	\label{tab:vlm}%
\end{table*}%

\noindent\underline{\textbf{Causal Intervention}}. After completing the aforementioned modeling, next, given a response $R_t$ generated by an LLM, we begin to analyze whether $R_t$ and $R$ are DAESO. 
Since $R_t$ does not have finely annotated explanations $E_{xp}$ like $R$, it's not easy to establish a causal chain as in Eq.~(\ref{eq:zk}). However, note that according to Section \ref{sec:testdata} and Fig.~\ref{fig:fill} illustrating the MLM task type, $R_t$ and $R$ differ only in the key text $\kappa$ part. Therefore, we consider intervening in the causal chain of $R$ in Eq.~(\ref{eq:zk}) with intervention $do(\kappa(R)\to\kappa(R_t))$ to approximate the chain that produces $R_t$, denoted as 
\begin{equation}
    \left(R_t^{(1)}, R_t^{(2)}, \cdots, R_t^{\left(\text{last}(R_t)\right)}\right),
\end{equation}
where $\text{last}(R_t)$ is the number of its nodes.
According to criterion (1), if $R_t$ and $R$ are DAESO, for Eq.~(\ref{eq:rc}), there also exists an $i^\prime \leq \text{last}(R_t)$ such that
\begin{equation}
    f(R_t^{(i^\prime)}) \to \mathcal{C}^{(j)}.
    \label{eq:t1}
\end{equation}
This implies that $R_t^{(i^\prime)}$ and $R^{(i)}$ both map to the same node $\mathcal{C}^{(j)}$ through the creative interpretation $f$, meaning that $R_t$ and $R$ share the same point of creativity. Next, from criterion (2), if $R_t$ and $R$ are DAESO, we have
\begin{equation}
    \Delta P\{R^{(\text{last}(R))}| do(\kappa(R)\to\kappa(R_t))\} \to 0, 
    \label{eq:t2}
\end{equation}
Where \(\Delta P\) denotes the change in probability. That is, after the causal chain of \(R\) is modified by replacing \(\kappa(R)\) with \(\kappa(R_t)\) and restructured, the last node \(R^{\text{last}(R)}\) that determines the functionality of these chains changes very little, i.e., $R^{\text{last}(R)} = R_t^{\text{last}(R_t)}$.

\subsubsection{Measuring DAESO by $\mathcal{E}_1$}

In this section, we provide the details of evaluator $\mathcal{E}_1$ by the modeling in Section \ref{sec:modeling}.
According to Section \ref{sec:modeling}, there are three steps to determine whether $R_t$ and $R$ are DAESO: (1) \textbf{Causal construction}. Establish the causal chain of $R$ and caption $\mathcal{C}$ based on $E_{xp}$ and $\mathcal{C}$; (2) \textbf{Causal Intervention}. Intervene in the causal chain of $R$ using $do(\kappa(R)\to\kappa(R_t))$ to build the causal chain of $R_t$; (3) \textbf{Judgment DAESO}. Determine whether Eq.~(\ref{eq:t1}) and Eq.~(\ref{eq:t2}) hold.

Considering the complexity of $E_{xp}$ and $\mathcal{C}$ across different samples, explicitly constructing the causal chain is highly challenging. If the chain is explicitly established, modeling the changes in chain nodes during key text interventions becomes difficult, making it hard to develop an effective algorithm to judge Eq.~(\ref{eq:t1}) and Eq.~(\ref{eq:t2}). To address this, in this paper, we propose using a powerful text-based LLM to map the causal chain between $R$ and caption $\mathcal{C}$ into the text space based on $E_{xp}$, $\mathcal{C}$, and a specially constructed prompt. This generates a long text to describe these causal chains, effectively establishing them. Leveraging the fault-tolerance and reorganization capabilities of LLMs, we semantically replace the key text $\kappa(R)$ with $\kappa(R_t)$ to reorganize the chain, then describe and judge Eq.~(\ref{eq:t1}) and Eq.~(\ref{eq:t2}) in language form, ultimately determining whether $R_t$ and $R$ are DAESO. In Section \ref{sec:ana}, we will validate the effectiveness of the proposed method for judging DAESO through a series of experiments. For details about the specific prompts, please refer to the supplementary materials.

\subsection{How to ask a Question $Q_t$ and answer it by $\mathcal{E}_2$?}
\label{sec:qa}

As mentioned in section \ref{sec:formulation}, the rethinking about spontaneous questioning is also a manifestation of one's own creativity \cite{sloane2003leader,thinking1970creativity}, which can visualize the process of achieving creativity in LLMs. Given the current input $I_t$ and an incorrect response $R_t$, LLMs utilize a question prompt $Q$, which includes a series of instructions related to questioning, such as system prompts, example prompts for in-context learning, and task-specific prompts. We require the LLM to propose a speculative question $Q_t$ about $R$, such as "Is it related to daily life?" or "Is it a type of appliance?" and so on, to help itself generate more human-like creative responses in the next round. Subsequently, for $Q_t$, we consider using a textual independent LLM, denoted as $\mathcal{E}_2$, to directly output a binary judgment $A_t$ containing Yes or No based on $R$. In Section \ref{sec:ana}, we find that selecting GPT-4o mini is suitable for this task. More details about the prompts are shown in the supplementary.

\begin{table*}[htbp]
  \centering
	
   \caption{The accuracy (\%) of choice questions and the NDCG (\%) of ranking questions on various \textbf{mutilmodal non-multilingual models} (English). See notations in Table \ref{tab:vlm}. We only consider I2T and T2T since English IT2T is not available due to cultural preference.  
   \vspace{-5pt}
}
    \resizebox*{1.00\linewidth}{!}{
    \setlength{\tabcolsep}{3mm} 
    \begin{tabular}{lr|cccccc|cccccc}
    \toprule
    \multirow{2}[4]{*}{Model} & \multirow{2}[4]{*}{Size} & \multicolumn{6}{c|}{Image to Text (I2T)}            & \multicolumn{6}{c}{Text to Text (T2T)} \\
\cmidrule{3-14}          &       & 2T1   & 3T1   & 4T1   & 5T2   & Rank  & Avg.$\qquad$  & 2T1   & 3T1   & 4T1   & 5T2   & Rank  & Avg.$\qquad$ \\
    \toprule
    InstructionBLIP~\cite{InstructBLIP} & 13B   & 19.8  & 13.7  & 15.5  & 1.1   & 65.5  & 23.1$\qquad$  & 22.3  & 16.0  & 17.0  & 0.7   & 59.5  & 23.1$\qquad$ \\
    mPLUG-Owl$_{\text{LLaMA2}}$~\cite{ye2023mplug} & 7B    & 22.3  & 12.7  & 15.0  & 4.2   & 59.9  & 22.8$\qquad$  & 24.2  & 13.7  & 12.6  & 3.1   & 59.2  & 22.6$\qquad$ \\
    Otter~\cite{li2023otter} & 7B    & 15.8  & 9.9   & 8.5   & 7.1   & 61.3  & 20.5$\qquad$  & 3.8   & 3.3   & 4.8   & 5.4   & 58.5  & 15.1$\qquad$ \\
    \midrule
    CogVLM-17B~\cite{wang2023cogvlm} & 7B    & 37.6  & 26.4  & 18.3  & 2.5   & 64.6  & 29.9$\qquad$  & 35.1  & 27.8  & 24.8  & 7.5   & 64.1  & 31.9$\qquad$ \\
    CogVLM-17B$_{+\text{AITv1 }}$ \cite{zhong2024let}& 7B    & 57.4  & 37.4  & 33.5  & 21.8  & 64.6  & 42.9$_{+13.1}$  & 55.4  & 46.5  & 26.4  & 18.2  & 64.4  & 42.2$_{+10.3}$ \\
    CogVLM-17B$_{+\text{CLoTv1 }}$ \cite{zhong2024let}& 7B    & \ul{66.9} & \ul{47.6} & \ul{43.4} & \ul{30.7} & \textbf{69.4} & \ul{51.6}{\color{black}$_{+21.7}$} & \ul{64.8} & \ul{52.9} & \ul{33.6} & \ul{21.8} & \ul{68.6} & \ul{48.3}{\color{black}$_{+16.4}$} \\
    CogVLM-17B$_{+\text{AITv2 }}$ (Ours)& 7B    & 59.2  & 39.1  & 35.5  & 23.8  & 65.2  & 44.6$_{+14.7}$  & 57.1  & 47.1  & 27.1  & 19.3  & 65.1  & 43.1$_{+11.2}$ \\
     CogVLM-17B$_{+\text{CLoTv2 }}$ (Ours)& 7B    & \textbf{68.4} & \textbf{49.8} & \textbf{46.4} & \textbf{32.5} & \ul{69.3} & \textbf{53.3}{\color{black}$_{+23.4}$} & \textbf{66.3} & \textbf{54.3} & \textbf{35.6} & \textbf{23.8} & \textbf{68.8} & \textbf{49.8}{\color{black}$_{+17.9}$} \\
    \bottomrule
    \end{tabular}%
   }
  \label{tab:non-multilingual}%
  \vspace{-10pt}
\end{table*}%

\begin{table}[t]
  \centering
     \caption{
  The accuracy (\%) of choice questions and the NDCG (\%) of ranking questions on various \textbf{large language models}. Here we use English T2T task for test.   See notations in Table \ref{tab:vlm}. 
  	\vspace{-1em}
 }
    \resizebox*{1.00\linewidth}{!}{
    \begin{tabular}{lr|ccccc}
    \toprule
    Model  & Size & 3T1   & 4T1   & 5T2   & Rank  & Avg. \\
    \midrule
    GPT-3.5~\cite{gpt4} & -     & 45.3  & 30.4  & 6.7   & 61.6  & 36.0 \\
    GPT-4~\cite{gpt4} & -     & 49.2  & 20.4  & 3.6   & 54.7  & 32.0 \\
    \midrule
    \multirow{3}[2]{*}{LLAMA2~\cite{touvron2023llama}} & 7B    & 18.9  & 13.5  & 1.1   & 60.4  & 23.5 \\
          & 13B   & 15.6  & 20.0  & 1.8   & 60.5  & 24.5 \\
          & 70B   & 27.8  & 16.1  & 3.8   & 62.0  & 27.4 \\
    \midrule
    \multirow{2}[2]{*}{Baichuan2~\cite{baichuan2023baichuan2}} & 7B    & 28.3  & 22.6  & 11.6  & 64.6  & 31.8 \\
          & 13B   & 21.7  & 18.3  & 8.9   & 61.5  & 27.6 \\
    \midrule
    \multirow{2}[2]{*}{Qwen~\cite{bai2023qwen}} & 7B    & 23.1  & 20.4  & 8.0   & 61.4  & 28.2 \\
          & 14B   & 27.4  & 22.2  & 12.3  & 59.5  & 30.3 \\
    \midrule
    ChatGLM3~\cite{du2022glm} & 6B    & 15.6  & 17.0  & 5.4   & 59.4  & 24.3 \\
    \midrule
    \multirow{2}[2]{*}{Vicuna-v1.5~\cite{vicuna2023}} & 7B    & 32.6  & 23.5  & 0.0   & 63.0  & 29.8 \\
          & 13B   & 30.2  & 23.0  & 2.7   & 62.2  & 29.5 \\
    \midrule
    Qwen-VL$_{+\text{CLoTv1}}$ \cite{zhong2024let}& 7B    & 51.7  & 32.3  & 24.8 & 65.0  & 43.4 \\
    CogVLM-17B$_{+\text{CLoTv1}}$ \cite{zhong2024let}& 7B    & 52.9 & \ul{33.6} & 21.8  & \ul{68.6} & 44.2 \\
     Qwen-VL$_{+\text{CLoTv2}}$ (ours)& 7B    & \textbf{53.9}  & 33.2  & \textbf{26.2} & 65.1  & \ul{44.6} \\
     CogVLM-17B$_{+\text{CLoTv2}}$ (ours)& 7B    & \ul{53.6} & \textbf{34.8} & \ul{23.1}  & \textbf{68.8} & \textbf{45.1} \\
    \bottomrule
    \end{tabular}%
   }

  \label{tab:llm}%
  \vspace{-10pt}
\end{table}%

\subsection{The Creativity Score $S_c$}
\label{sec:sc}

As mentioned in section \ref{sec:lotbench_main}, LoTbench aims to explore how many rounds of creative thinking are required for LLMs to achieve HHCRs, with fewer rounds indicating statistically higher creativity. Therefore, we believe the creativity score \( S_c \) should meet at least two requirements for the set of the number of rounds \(\mathbf{r} = [t_r^{(1)}, t_r^{(2)}...,t_r^{(m)}]\) with $m$ times repeated evaluation: (1) As \(\min(\mathbf{r}) \to \infty\), \( S_c \to 0 \), meaning that if an LLM has not reached the creativity level of HHCR after a sufficiently large number of rounds, its contribution to creativity in the current sample tends to zero; (2) \( S_c \) should be inversely proportional to the round, meaning that the faster the LLM reaches HHCR, the more creative it is considered to be. Thus, we propose the following formula to define \( S_c \).
\begin{equation}
    S_c = \frac{1}{mn}\sum\nolimits_{j=1}^m\sum\nolimits_{r=1}^n \beta_c\exp[-\alpha_c\cdot t_r^{(j)} ],
\end{equation}
where $n$ represents the number of test samples in LoTbench, while $\beta_c$ and $\alpha_c$ are hyperparameters, and set to 1.0 and 0.2 respectively in this paper. The $m$ denotes the number of rounds for repeating independent tests on a single sample. The rationale behind conducting multiple experiments is to reduce the errors in evaluator judgments as shown in Section~\ref{sec:ana} and provide more opportunities for LLMs to engage in creative thinking, as creative responses are not always produced \cite{zhong2024let}. Additionally, considering the cost of inference, we set $m=3$.

\section{Experiments under Standard Evaluation}
\label{sec:exp}

In this section, we explore the creativity of different LLMs through standard evaluation shown in Section \ref{sec:std}, while testing the creative response generation capability of CLoTv2 proposed in Section \ref{sec:syn}.
Noticing that we considered setting the condition in the instruction to empty $\phi$ in both the associable instruction tuning and Explorative Self-Refinement stages, this ensures that the LLM can generate creative responses without specific conditions, facilitating practical use of the model without the need to set conditions. Therefore, under the settings of CLoTv2, after training, the LLM can directly perform model inference through the instruction shown in Fig.~\ref{fig:template}.

\subsection{Evaluation by Choice and Ranking Questions}
\label{sec:stdeval}

\noindent\textbf{Evaluation on Multimodal Multilingual LLMs.} We plug our associable instruction tuning (AITv2) and our CLoTv2 into the advanced open-source multimodal multilingual model Qwen-VL~\cite{Qwen-VL} to obtain Qwen-VL$_{+\text{AITv2}}$ and Qwen-VL$_{+\text{CLoTv2}}$, respectively.    
Table~\ref{tab:vlm} shows that,  on three tasks (IT2T, I2T and T2T)  which include English, Chinese and Japanese questions,  Qwen-VL achieves the best LoT performance among all open-source baselines in most cases. In comparison, Qwen-VL$_{+\text{AITv2}}$ achieves a noticeable improvement on the advanced Qwen with average accuracy enhancements of 7.4\%, 7.6\%, and 5.3\% on the three tasks, respectively.
Importantly, Qwen-VL$_{+\text{CLoTv2}}$ further enhances Qwen-VL, showing improvements of 9.6\%, 11.7\%, and 9.1\% in accuracy across these tasks.
These results demonstrate the efficacy of the two stages in CLoTv2, i.e.,  associable instruction tuning and explorative self-refinement.

\begin{figure}[t]
  \centering
 \includegraphics[width=0.99
\linewidth]{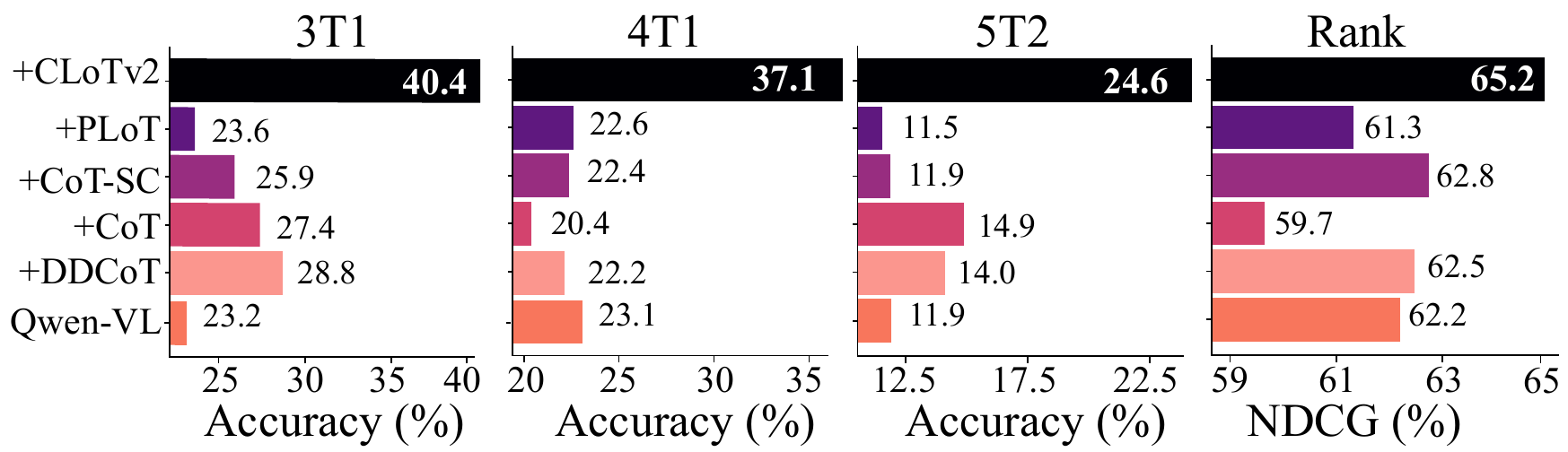}
 \vspace{-10pt}
  \caption{
  	\hzz{The accuracy (\%) of choice questions and the NDCG (\%) of ranking questions on our CLoT and various reasoning frameworks. The baseline is Qwen-VL on multilingual I2T task.  For $m$T$n$ choice questions,  one needs to select $n$ correct answers from   $m$  options.}
  	}
\label{fig:reasoning}
\end{figure}

\noindent\textbf{Evaluation on Multimodal Non-multilingual LLMs.} Here we integrate our CLoTv2 with the  advanced multimodal non-multilingual model, CogVLM-17B~\cite{wang2023cogvlm}, and evaluate  it on the English I2T and T2T tasks. 
Table \ref{tab:non-multilingual} shows that CogVLM-17B$_{+\text{AITv2}}$ achieves remarkable improvements over the standard CogVLM-17B, and CogVLM-17B$_{+\text{CLoTv2}}$ consistently demonstrates significantly superior performance compared to CogVLM-17B.

\noindent\textbf{Evaluation on Single-Modal LLMs.}   Now we test LLMs that can handle only pure texts, using the English T2T task for evaluation.  
Table~\ref{tab:llm} also indicates the insufficient   LoT ability within existing LLMs, ranging from small to large models.  
Fortunately, our CLoTv2 significantly improves the LoT ability of these LLMs,
as demonstrated by the notable improvement in accuracy.

\noindent\textbf{Comparison with CoT-alike Reasoning Frameworks.} We also find that existing reasoning frameworks are not as effective as CLoTv2 in enhancing LoT ability.  Fig. \ref{fig:reasoning} \hzz{compares CLoTv2 with CoT~\cite{kojima2022large,wei2022chain}, CoT-SC~\cite{wang2022self}, DDCoT~\cite{zheng2023ddcot}, and prompted-based LoT (PLoT) with the prompt ``let's think outside the box".} The results reveal that CoT-alike frameworks do not enhance LoT performance of LLMs, while CLoT framework demonstrates the ability to consistently enhance LLMs. 

Our experiments and analysis reveal that, unlike CoT-based methods, LoT cannot be directly achieved by prompting alone. 
This is because the inherent reasoning capabilities and extensive knowledge of LLMs are not sufficient to enable LoT ability.
However, when trained with our proposed CLoTv2 method, LLMs can effectively engage in a range of creative tasks. Additionally, the use of specific prompting techniques can enhance the LoT ability of CLoTv2-trained LLMs. These findings suggest that LoT could potentially be considered an additional general reasoning ability for LLMs
that is not contained in current LLMs or we may need to use more advanced methods to stimulate their LoT.

\begin{figure}[t]
  \centering
 \includegraphics[width=0.9\linewidth]{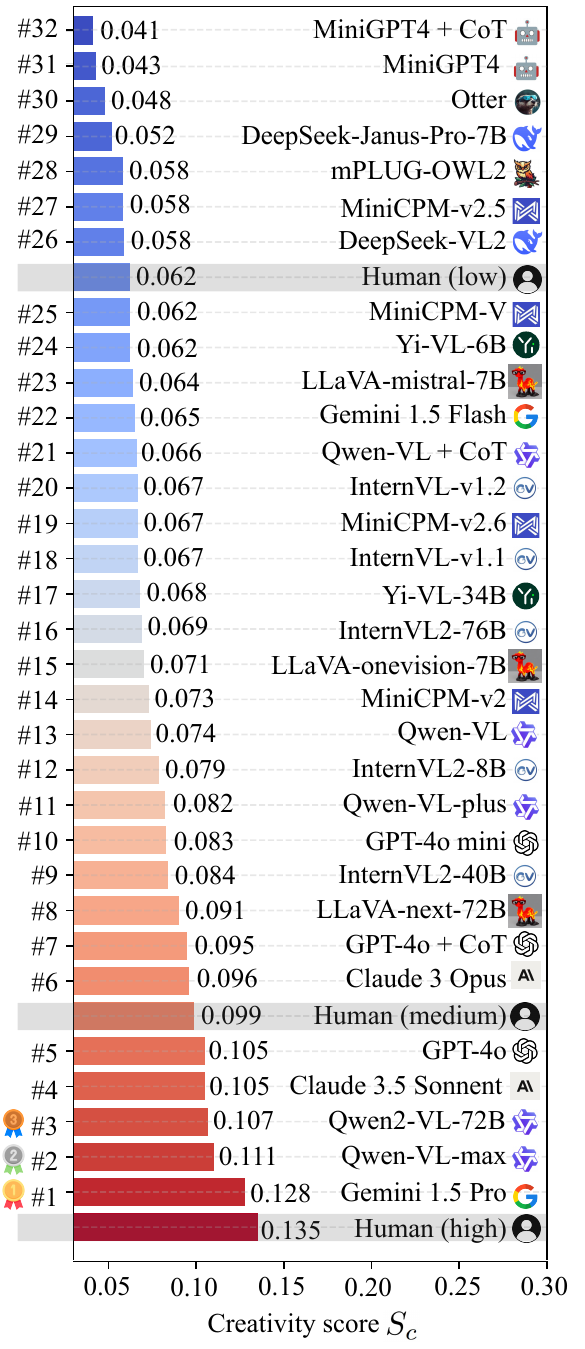}
 \vspace{-10pt}
  \caption{The ranking results of LLM's creativity by LoTbench. 
}
\label{fig:lotresult22}
\vspace{-10pt}
\end{figure}

\subsection{Experiments under LoTbench}
\label{sec:result}

In this section, we assess the creativity of various multimodal LLMs using LoTbench. To better understand the creativity score $S_c$, we introduced 21 human subjects aged between 13 and 44, testing only the samples in LoTbench for languages they are proficient in. Ultimately, we divided their $S_c$ into three equal groups based on their $S_c$ rankings, with each group containing 9 individuals, and calculated the average $S_c$ for reference, naming them human (high), human (medium), and human (low). 
	\begin{figure}[t]
  \centering
 \includegraphics[width=0.99\linewidth]{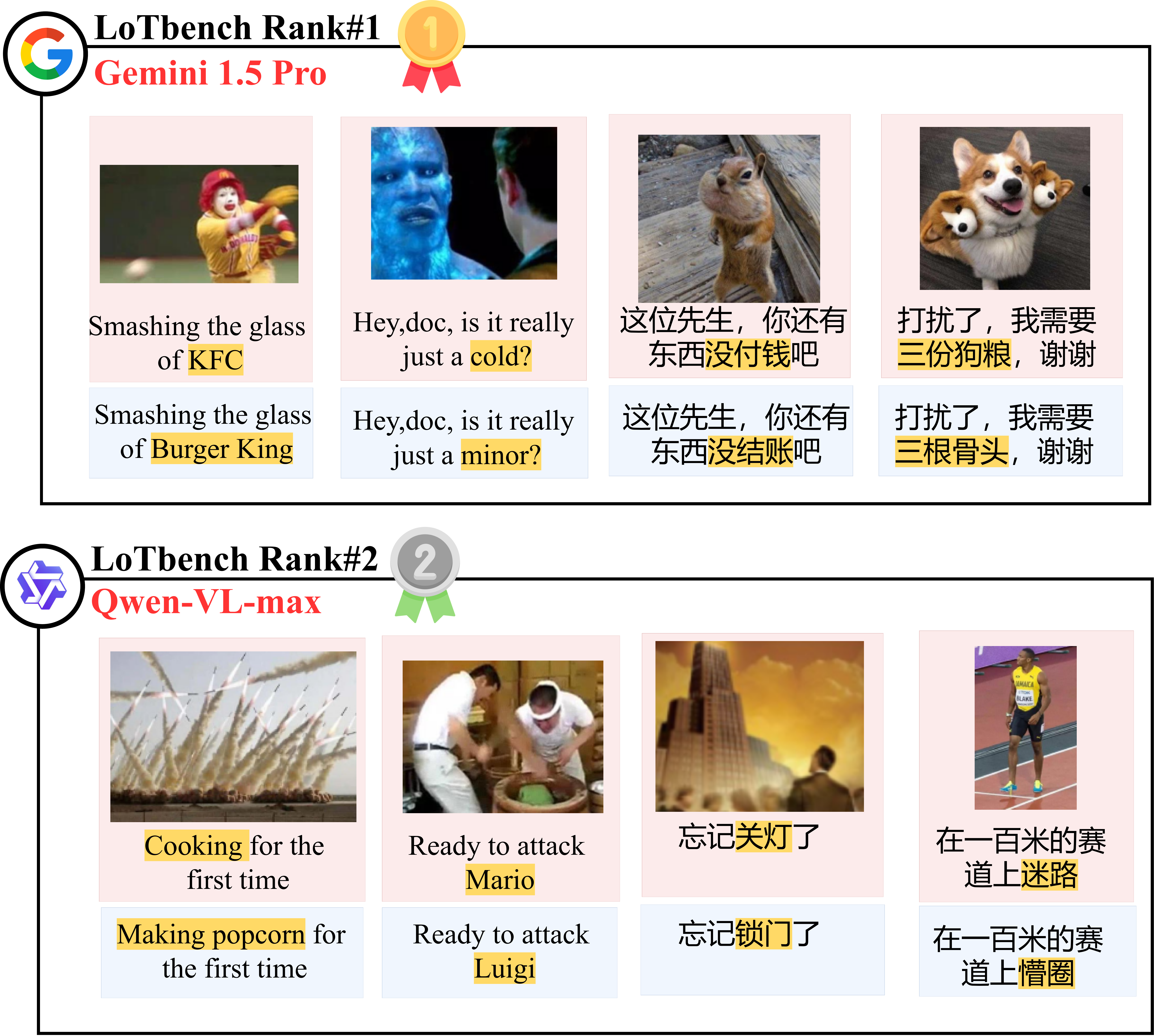}
 \vspace{-10pt}
  \caption{Specific creative responses. We visualize the outputs of the two best-performing LLMs shown in Fig.~\ref{fig:lotresult22}. The red boxes indicate the original HHCRs, while the blue boxes are these LLMs' final outputs.
}
\label{fig:yq}
\end{figure}

	\begin{figure}[t]
  \centering
 \includegraphics[width=0.9\linewidth]{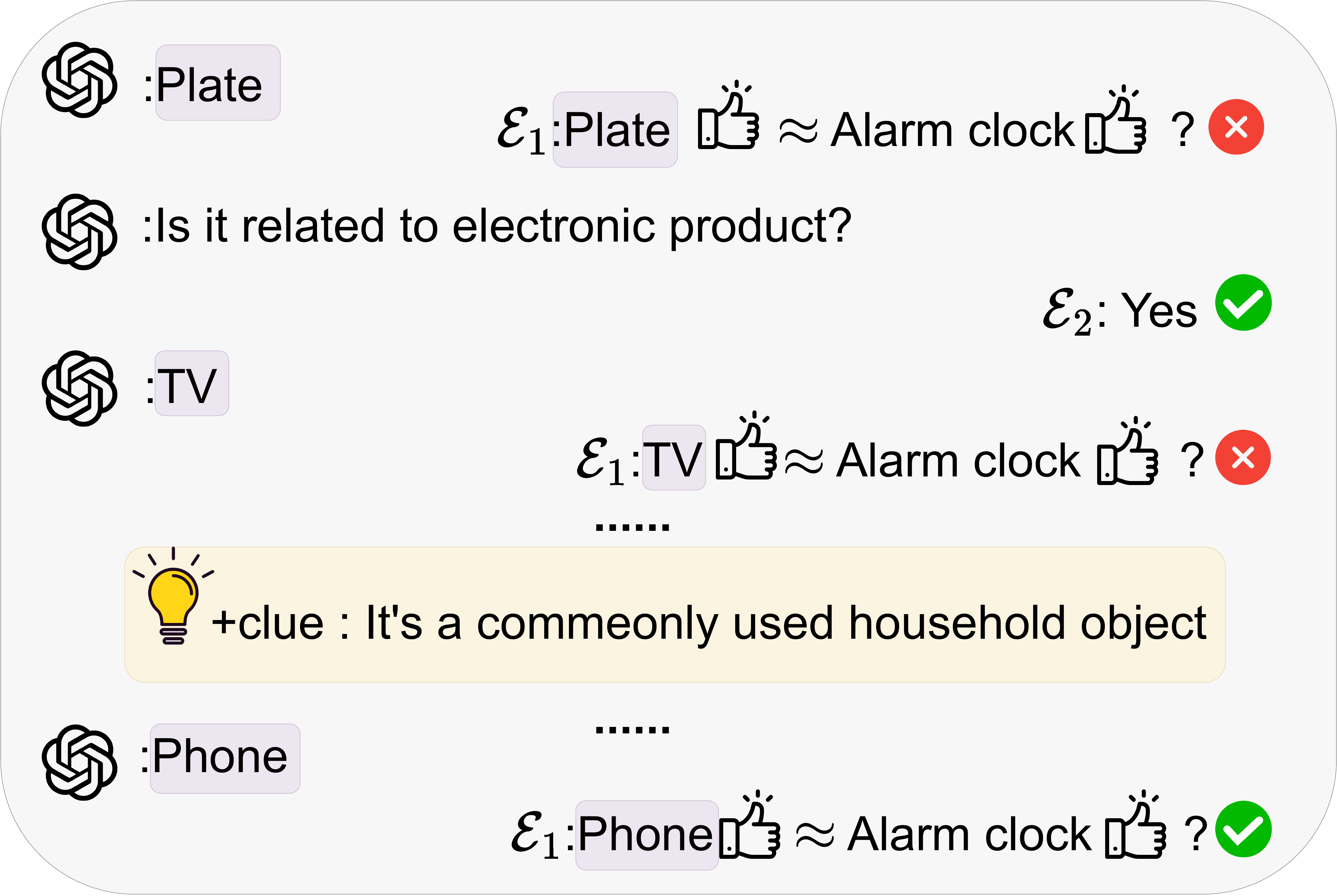}
 \vspace{-10pt}
  \caption{Example of visualization for creative thinking in LoTbench.
}
\label{fig:vis}
\vspace{-10pt}
\end{figure}

From the results in Fig.~\ref{fig:lotresult22}, most LLMs do not exhibit high creativity in the  LoTbench scenario, but the gap between their creativity and the average level of human participants is not particularly large. On one hand, generating creativity is inherently difficult for both humans and LLMs, which may be a primary reason for the current scarcity of creativity data. On the other hand, the testing process in LoTbench requires multiple rounds of interaction. We found that, after several interactions, human subjects of all levels often experience pauses, such as hesitations about how to respond. This represents a disadvantage for humans in long-term interactive evaluations. However, LLMs, which usually are based on next token prediction~\cite{zhong2024let,bai2023qwen,vicuna2023,zhong2023adapter}, do not face this issue and can continuously generate responses. Therefore, LoTbench, designed specifically for LLMs, may not fully capture the average creativity level of humans during testing and should be considered as a reference. In this sense, the creativity levels of currently strong LLMs and human subjects are quite similar. Furthermore, since most LLMs generate responses based on next token prediction, if they can be sufficiently stimulated for creativity, they have the potential to produce a vast number of responses continuously, some of which may contain valuable and highly creative ideas. This could be an important direction for future scientific advancement. \hzz{For all LLM in Fig.~\ref{fig:lotresult22},   the average number of rounds to complete the test is 13.65, with the best-performing Gemini 1.5 Pro achieving an average of 12.44 rounds. According to our setup in Section~\ref{sec:testdata}, the upper limit for the number of rounds per sample is 15, which intuitively reflects the relatively low creativity level of current LLMs. This limited capability results in an average forward inference cost that is over more than 10 times higher in the current LoTbench testing paradigm.}. Next, in Fig.~\ref{fig:yq}, we  illustrate specific response examples for the first and second place winners in Fig.~\ref{fig:lotresult22}.

	Moreover, in Fig.~\ref{fig:vis}, we take Fig.~\ref{fig:fill} example 1 as a case study to demonstrate the entire process of GPT4o mini undergoing the LoTBench assessment. This process visualizes the entire thought process of GPT4o mini, providing an interpretable observation for humans to understand the creative generation process of LLMs.

\section{Analysis}
\label{sec:ana}


\begin{figure}[t]
  \centering
 \includegraphics[width=0.99\linewidth]{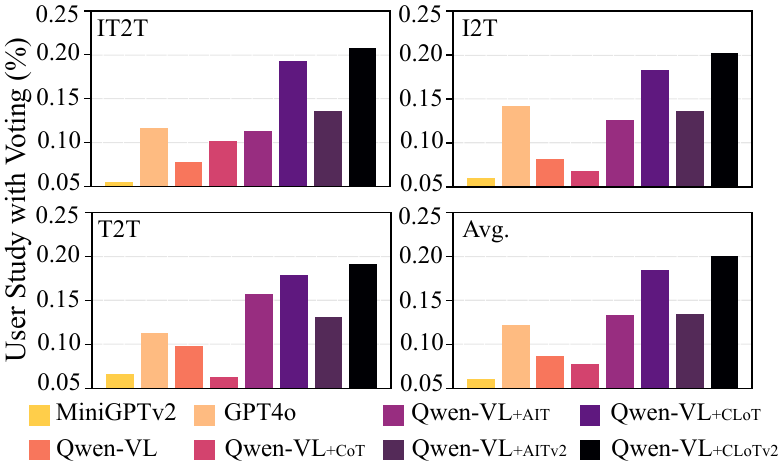}
  \caption{\hzz{User study with voting (\%) for Oogiri-style creative responses by different models and improved methods. ``Avg." denotes the avervage voting of three types of Oogiri game results.} }
\label{fig:user-study}
\end{figure}

\subsection{Ohter Types of Evaluation}

In Section \ref{sec:lotbench_main}, we shown that the truly reasonable creativity evaluation should assess the "measure the creativity level of LLM" rather than "recognize the creativity from LLM."
Actually, to evaluate the "measure the creativity level of LLM", there are also some previous methods like human evaluation and LLM-as-a-Judge~\cite{liu2023g,mao2023gpteval,ge2023mllm,zheng2023judging}. In this section, we consider these two kinds of evaluations for LLM's creativity, and show the necessity of LoTbench.

\subsubsection{Human Evaluation for LLMs'  Creativity}
\label{sec:human}
We conduct a user preference study to test creativity of LLMs. Here we select eight LLMs to generate responses for a total of twenty-one questions across three tasks (IT2T, I2T and T2T). We use choice questions, and ask users to choose the creative and humorous responses they think. Fig. \ref{fig:user-study} summarizes the statistical analysis of 56 valid surveys. The results indicate a strong user preference for the enhanced outputs from both the associable instruction tuning and explorative self-refinement stages across all three tasks, highlighting the effectiveness of our proposed method for synthesizing HHCRs. See more details in Appendix of the conference version~\cite{zhong2024let}. 
The human evaluation provides the reasonable evaluation since it assesses whether the generated responses directly align with human creativity. However, it has clear drawbacks: it is unsustainable and requires additional manpower, leading to high costs when evaluating the creativity of new models. Additionally, for the new LLM under test, there might be fairness issues if the participants voting each time are different.

\subsubsection{LLM-as-a-Judge for LLMs'  Creativity}
\label{sec:asjudge}

Moreover, following analysis through the Oogiri game, we find that directly using LLM-as-a-Judge might also struggle to accurately assess creativity of LLM.

\begin{figure}[t]
  \centering
 \includegraphics[width=0.99\linewidth]{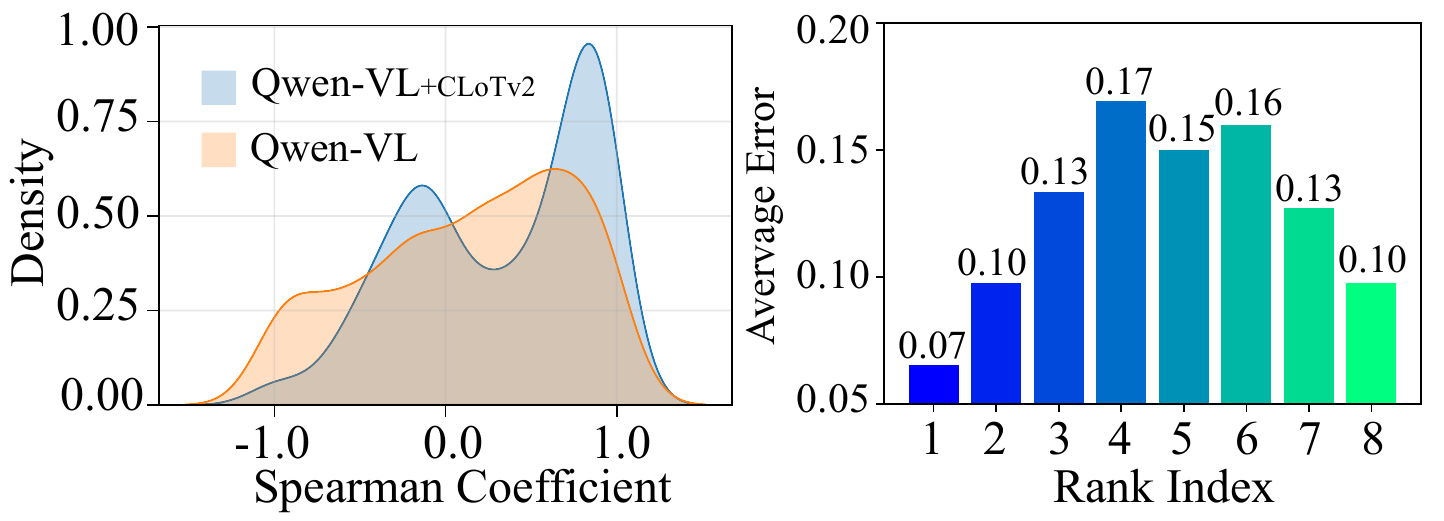}
 \vspace{-0.2cm}
  \caption{The comparison between human preference and  LLM-as-a-Judge on Oogiri game. (Left) Distribution of Spearman's rank correlation between different LLM and user study results. (Right) The ranking error of Qwen-VL$_{+\text{CLoTv2 }}$ under different rank indices.}
\label{fig:llmeva}
\vspace{-10pt}
\end{figure}

Specifically, we consider the following settings to explore the relationship between LLM-as-a-Judge and human preferences in Fig.~\ref{fig:user-study}. Using the setup shown in Fig.~\ref{fig:user-study}, we randomly select 5 responses out of 8 generated for each sample and construct a ranking based on human preferences as the ground truth. Then, we have Qwen-VL and Qwen-VL$_{+\text{CLoTv2 }}$ rank the 5 responses as well, comparing their rankings to the ground truth using the Spearman (SP) rank correlation coefficient~\cite{arsov2019measure}. A higher SP score indicates a ranking closer to the human preference; a lower score indicates less similarity. As shown in Fig.~\ref{fig:llmeva} (Left), we observe that while the original Qwen-VL has some ability to align with human-recognized creativity, it is nearly unusable in practice since a large number of SP values are concentrated around zero, and even in the negative region. In contrast, the enhanced Qwen-VL$_{+\text{CLoTv2 }}$ significantly improves the LLM's judgment of responses. These observations are consistent with the standard evaluation results in Section \ref{sec:stdeval}. However, since SP scores for Qwen-VL$_{+\text{CLoTv2 }}$ occasionally fall below 0.0, it indicates that this model is not fully reliable for judging or precisely scoring arbitrary responses. 

Next, we further analyze Qwen-VL$_{+\text{CLoTv2 }}$'s ranking accuracy by counting the ranking errors across different rank indices. Fig.~\ref{fig:llmeva} (Right) shows the error distribution across these indices. We observe that Qwen-VL$_{+\text{CLoTv2 }}$ has fewer errors at the highest and lowest ranks, performing best at identifying the highest- and lowest-quality responses, while its performance is more ambiguous with mid-range quality responses. This suggests that although Qwen-VL$_{+\text{CLoTv2 }}$ may struggle to give fine-grained scores or make fully accurate judgments for any sample, it remains viable as a data filtering tool as discussed in Section \ref{sec:syn}.

\begin{figure*}[t]
  \centering
 \includegraphics[width=0.99\linewidth]{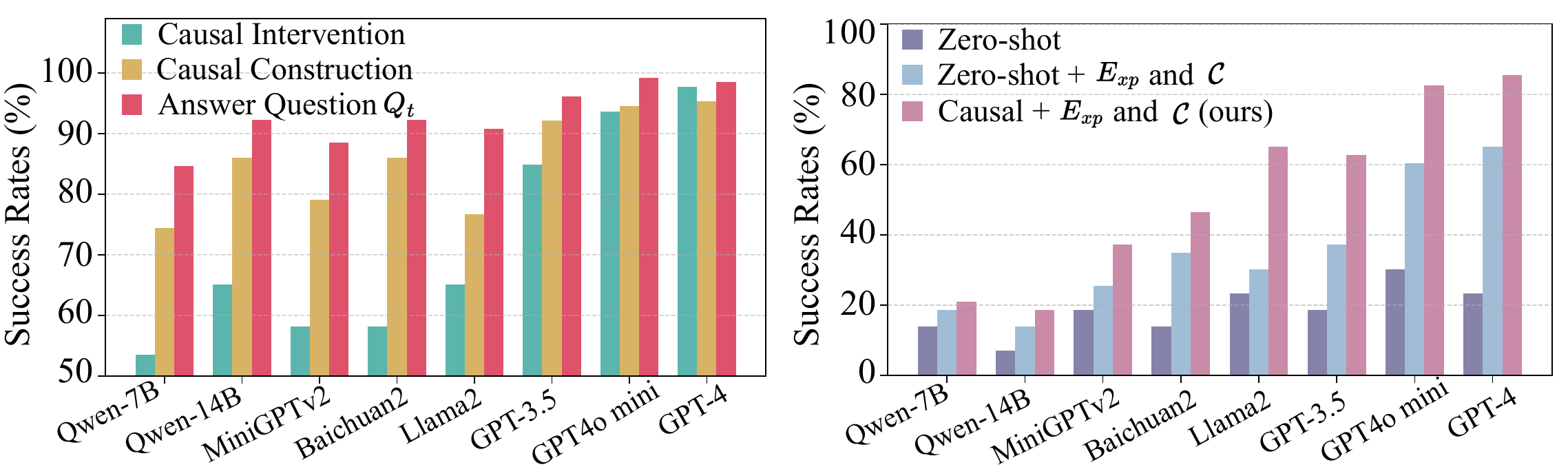}
 \vspace{-10pt}
  \caption{Analysis of the discriminative abilities of $\mathcal{E}_1$ and $\mathcal{E}_2$. (Left) We examined the success rates (\%) of various advanced LLMs in performing causal construction and causal intervention as described in Section \ref{sec:modeling}. We also tested whether these LLMs could effectively engage the target LLM in interactive responses as outlined in Section \ref{sec:qa}. (Right) We analyzed the accuracy (\%) in determining whether $R_t$ and $R$ are DAESO under different settings.}
\label{fig:evalcas}
\vspace{-10pt}
\end{figure*}

\begin{figure*}[t]
  \centering
 \includegraphics[width=0.99\linewidth]{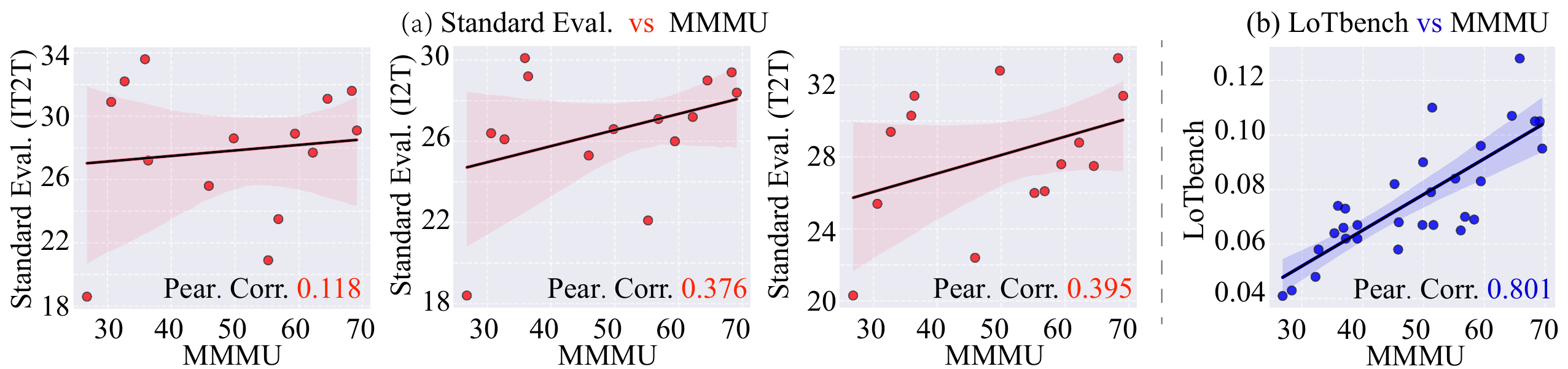}
 \vspace{-10pt}
  \caption{The correlation analysis between LLM cognition benchmark MMMU and the proposed creatvity benchmark. (a) The correlation between MMMU and average results from different types of standard evaluation in Table~\ref{tab:vlm}.  (b) The correlation between MMMU and LoTbench. blue dot \coloredcircle[blue]  and red dot \coloredcircle[red]   denote the multimodal LLM to be tested in LoTbench and standard evaluation, respectively.
}
\label{fig:lotresult}
\vspace{-10pt}
\end{figure*}

In summary, due to the limitations of different evaluation methods mentioned above, we propose LoTbench in this paper is necessary. Instead of directly scoring LLM responses, LoTbench estimates creativity by measuring the average cost required for the target LLM to achieve carefully designed HHCRs through interaction with some specific LLMs. Lower costs indicate higher creativity. This approach maintains the automation advantages of LLM-as-a-Judge based methods through the involvement of a specific LLM, while considering the average distance to HHCRs ensures alignment with human preferences. Additionally, the carefully designed HHCRs and interactive approach help mitigate the risk of information leakage in standard evaluation and improve the interpretability.

\subsection{The Effectiveness of $\mathcal{E}_1$}
To validate this LLM-based DAESO judgment method proposed in Section \ref{sec:daeso}, we used a validation set of 43 DAESO samples collected during the construction of LoTbench to perform tests from multiple perspectives. First, we conducted experiments on causal chain construction and intervention using Qwen-7B, Qwen-14B, MiniGPTv2, Baichuan2, Llama2, GPT-3.5, GPT4o mini and GPT-4 with this validation set, manually inspecting each result for accuracy. As shown in Fig.~\ref{fig:evalcas} (Left), we can find significant performance differences in causal chain construction and intervention among these LLMs, with GPT series outperforming the others.  Moreover, in Fig.~\ref{fig:evalcas} (Right), we considered three settings: (1) Using only a prompt to let the LLM directly zero-shot determine whether $R_t$ and $R$ are DAESO; (2) Zero-shot DAESO judgment with detailed $E_{xp}$ and $\mathcal{C}$ provided; (3) Our proposed method of causal chain modeling in text space based on $E_{xp}$ and $\mathcal{C}$. From the results, we can see that $E_{xp}$ and $\mathcal{C}$ are crucial—without them, all LLMs performed poorly, almost guessing randomly. When this information is provided, modeling and intervening in the causal chain resulted in better judgment outcomes. Therefore, based on these experimental results, to ensure the accuracy of LoTbench and minimize reasoning costs, like API fee, we adopt GPT-4o mini as $\mathcal{E}_1$ in the actual evaluation of LoTbench and provided detailed $E_{xp}$ for each sample.

\subsection{The Effectiveness of $\mathcal{E}_2$}
In Section \ref{sec:qa}, we need a powerful text-based LLM as evaluator $\mathcal{E}_2$ to accurately respond to the tester's spontaneous questioning $Q_t$ with a precise answer $A_t$.
In this section, we find that selecting GPT-4o mini is suitable for this task as shown in Fig.~\ref{fig:evalcas} (Left).
Specifically, during the construction of the LoTbench test set, we also collect a simple validation set of 130 examples to test whether various LLMs have the ability to provide judgments for $Q_t$. Fig.~\ref{fig:evalcas} (Left) shows the judgment results of different LLMs, where most LLMs can achieve an accuracy rate of over 80\%, and GPT-4o mini not only has a relatively low reasoning cost but also an accuracy rate of up to 98\%. Therefore, we choose it as the judge $\mathcal{E}_2$.

\subsection{The Correlation with Cognition Benchmark}

(2) The creativity assessment results from LoTbench align more closely with current human cognitive theories~\cite{martinsen1994effect,martinsen1993insight,kaufmann1979explorer,runco1995cognition,mednick1962associative}. 
In this Section, we compare the evaluation results of the well-known and comprehensive cognitive ability assessment MMMU with those of our proposed LoTbench in Fig.~\ref{fig:lotresult} (b). We found that, although MMMU focuses on cognitive abilities while LoTbench emphasizes the creativity of LLMs, there is a significant strong correlation between their results. This indicates that the evaluation of LoTbench aligns with current human cognitive theories and reflects human-like creativity, suggesting that cognition forms the basis of creativity. The components of perception, knowledge, and reasoning enable the recognition and connection of different concepts through creative thinking, which is key to early creativity.
However, it is important to note that in Fig.~\ref{fig:lotresult22}, we also observe that methods like CoT, which enhance LLM logical reasoning abilities, do not always improve $S_c$. This is consistent with the observations in Section \ref{sec:exp}. It suggests that creativity requires higher demands on cognitive abilities, and merely enhancing logical reasoning is insufficient to consistently boost creativity.
Furthermore, we also analyze the standard evaluation shown in Section \ref{sec:exp} in the same way, and the results, shown in Fig.~\ref{fig:lotresult} (a), indicate that the relationship between "creativity"  by standard evaluation and LLM cognitive ability is not very significant. Of course, this does not imply that standard evaluation is an incorrect benchmark, and it indeed helps humans understand LLM creativity from different perspectives and provides preliminary quantitative results.

\subsection{Limitation}
\hzz{While the proposed LoTbench offers intuitive and user-friendly features, it does have certain limitations. For instance, as a multi-turn interactive benchmark, it may not be suitable for evaluating all types of subjects.
For example, as mentioned in Section \ref{sec:result}, humans are not particularly adept at long-term interactions. Similarly, CLoTv2 faces this issue as well. Due to the multi-turn tuning described in Section \ref{sec:lotbench_main}, while its creativity shows some improvement in standard evaluations, other abilities may be slightly diminished~\cite{zhai2023investigating}, especially the ability to follow context. This limitation prevents it from being assessed using LoTbench.} Currently, LoTbench is primarily designed for evaluating the creativity of general LLMs. In the future, we need to explore better methods to stimulate LLM creativity using LoTbench while minimizing the catastrophic forgetting~\cite{zhai2023investigating} of general capabilities. On the other hand, LoTbench constructs creativity scores by estimating the average cost for the target LLM to achieve certain HHCRs through interactive methods. While this definition of LLM creativity facilitates benchmark design, it is not the only definition, and the concept of creativity still lacks a clear consensus~\cite{elgammal2015quantifying,egan2017developing}. In the future, the community should explore more advanced definitions of creativity to address potential risks associated with LoTbench.

\vspace{-0.10cm}
\section{Conclusion}
\label{sec:ana}
\hzz{This paper investigates creativity in LLMs and provides an in-depth analysis of their Leap-of-Thought (LoT) abilities through the Oogiri game. In particular, given some inherent issues, like information leakage, in current assessments of LLM creativity, we introduce a novel interactive benchmark, LoTbench, to effectively evaluate LLM creativity. Our findings reveal that while LLMs exhibit limited creativity, the gap between LLM and human creativity is not significant. 
Additionally, 
we find a strong correlation between LoTbench results and MMMU, a comprehensive benchmark for multimodal LLM cognition. This suggests that LoTbench aligns with human cognitive theories, capturing human-like creativity and emphasizing cognition as a foundational element in the early stages of creativity, enabling the integration of diverse concepts.}

	
	%




	\section*{Acknowledgments}
    This work was supported by National Science and Technology Major Project (No.2021ZD0111601), National Natural Science Foundation of China under Grants No. 623B2099 and 62325605, Guangdong Basic and Applied Basic Research Foundation (No.2023A1515011374), and Guangzhou Science and Technology Program (No.2024A04J6365).
Pan Zhou was supported by the Singapore Ministry of Education (MOE) Academic Research Fund (AcRF) Tier 1 grants (project ID: 23-SIS-SMU-028 and 23-SIS-SMU-070).


	\ifCLASSOPTIONcaptionsoff
	\newpage
	\fi

	
	
	
	\bibliographystyle{IEEEtran}
	\typeout{}
	\bibliography{ref2.bib}
	
	%

	
	
	
	%
	\clearpage
	\onecolumn
	\section*{Appendix A: The details of $I_t$ and generation of $R_t$}

\begin{mdframed}[backgroundcolor=gray!8]
\begin{minipage}{\linewidth}

\sethlcolor{prompt}\hl{Now, please help me complete a task involving filling in a blank with a humorous and creative . All the input information is provided in the INPUT section, which includes:}

	1.	\sethlcolor{condition}\hl{IMAGE}: \sethlcolor{prompt}\hl{A given image.}
    
	2.	\sethlcolor{condition}\hl{IMAGE CAPTION}: \sethlcolor{prompt}\hl{A detailed description of the given image.}
    
	3.	\sethlcolor{condition}\hl{RESPONSE}: \sethlcolor{prompt}\hl{A sentence with a blank  to be filled. You need to complete the  part based on all the INPUT information.}
    
	4.	\sethlcolor{condition}\hl{TIPS}: \sethlcolor{prompt}\hl{Some guidelines for the task, which include:}
    
	•	\sethlcolor{condition}\hl{Q\&A}: \sethlcolor{prompt}\hl{Questions and answers related to . Carefully analyze and follow these hints to generate a creative and humorous .}
    
	• \sethlcolor{condition}\hl{CLUE}: \sethlcolor{prompt}\hl{Descriptive hints about  Understand and adhere to these clues to generate a creative and humorous .}
    
	•	\sethlcolor{condition}\hl{WRONG-ANS}: \sethlcolor{prompt}\hl{Examples of  that are neither innovative nor humorous. Avoid completing the blank with similar content.
    }

\sethlcolor{response}\hl{Based on the provided INPUT information and image, use divergent thinking to complete the  part. Ensure that the final RESPONSE pairs well with the IMAGE, making it witty, humorous, and creative. Output the result in the specified OUTPUT format.
}\\

Here are some examples:
\\

\sethlcolor{image}\hl{Example1}:
\\

INPUT: {
"IMAGE": ,

"IMAGE CAPTION": "A soldier holding a big knife, staring angrily ahead, seems very angry",

"RESPONSE": "$\langle$WORD$\rangle$ After reading my paper...",

"TIPS": {

\quad "WRONG-ANS ($\langle$WORD$\rangle$ is not the following content)": {

\quad\quad 1: "Programmer",

\quad\quad 2: "Mountain climber"},

"SYSTEM CLUE": {

\quad "CLUE1": "$\langle$WORD$\rangle$ is a kind of person",

\quad "CLUE2": "This kind of person has high knowledge"},

"Q\&A (OUTPUT should not be repeated with Q\&A)": 

{
1: {
"Q1": "Is it related to the soldier?",

\quad "A1": "No"},

2: {
"Q2": "Is it related to the school?",

\quad "A2": "Yes"
}
}
}
}
\\

OUTPUT: 

{
"$\langle$WORD$\rangle$": "Tutor",

"RESPONSE": "After the tutor read my paper..."
}\\

......

\sethlcolor{image}\hl{Example2}: ......

\sethlcolor{image}\hl{Example3}: ......
\\

\sethlcolor{prompt}\hl{Referring to the example above, please use the latest INPUT information provided below and the accompanying image to creatively and humorously complete the $\langle$WORD$\rangle$. Ensure that the supplemented RESPONSE matches the provided IMAGE, making it witty, imaginative, and engaging. Format the result strictly according to the example shown in the OUTPUT.}\\

\sethlcolor{response}\hl{
INPUT: 

   "IMAGE": "$\langle$image$\rangle$",
   
   "IMAGE CAPTION": "$\langle$caption$\rangle$",
   
   "RESPONSE": "$\langle$response$\rangle$",
   
   "TIPS": $\langle$tips$\rangle$

OUTPUT: 
}

\end{minipage}
\end{mdframed}    

\newpage

\section*{Appendix B: The details of rethinking and generating $Q_t$}

\begin{mdframed}[backgroundcolor=gray!8]
\begin{minipage}{\linewidth}

\sethlcolor{prompt}\hl{Given an image IMAGE and its detailed description IMAGE CAPTION. It is known that this IMAGE has a very humorous and creative caption RESPONSE.

Now there is a task of looking at the image to complete the caption, that is, to generate a humorous and creative $\langle$WORD$\rangle$. All the input information is in INPUT, they are}

\quad 1.\quad \sethlcolor{condition}\hl{IMAGE}: \sethlcolor{prompt}\hl{a given image}

\quad 2.\quad \sethlcolor{condition}\hl{IMAGE CAPTION}: \sethlcolor{prompt}\hl{a detailed description of the given image IMAGE}

\quad 3.\quad \sethlcolor{condition}\hl{RESPONSE}: \sethlcolor{prompt}\hl{a text of an IMAGE with the content $\langle$WORD$\rangle$ to be completed, you need to complete the $\langle$WORD$\rangle$ part according to IMAGE and IMAGE CAPTION}

\quad 4.\quad \sethlcolor{condition}\hl{Q\&A}: \sethlcolor{prompt}\hl{some known queries and corresponding answers about $\langle$WORD$\rangle$}

\quad 5.\quad \sethlcolor{condition}\hl{CLUE}: \sethlcolor{prompt}\hl{some descriptive hints related to $\langle$WORD$\rangle$}

\quad 6.\quad \sethlcolor{condition}\hl{WRONG-ANS}: \sethlcolor{prompt}\hl{some innovative and humorous $\langle$WORD$\rangle$, you should not complete similar content}

\sethlcolor{prompt}\hl{In order to better complete the $\langle$WORD$\rangle$ in RESPONSE, so that the combination of IMAGE and RESPONSE is very humorous and creative, you can first use divergent thinking to ask a general question (that is, a question with the answer of Yes or No) for possible $\langle$WORD$\rangle$ to help the generation of $\langle$WORD$\rangle$.}\\

Here are some examples:\\

\sethlcolor{image}\hl{Example1}:\\

INPUT: {

"IMAGE": ,

"IMAGE CAPTION": "A soldier holding a big knife, staring angrily ahead, seems very angry",

"RESPONSE": "$\langle$WORD$\rangle$ After reading my paper...",

"TIPS": {

\quad "WRONG-ANS ($\langle$WORD$\rangle$ is not the following content)": {

\quad \quad 1: "Programmer",

\quad \quad 2: "Mountain climber"},

"SYSTEM CLUE": {

\quad "CLUE1": "$\langle$WORD$\rangle$ is a kind of person",

\quad "CLUE2": "This kind of person has high knowledge"},

"Q\&A (OUTPUT should not be repeated with Q\&A)": {

\quad 1: {

"Q1": "Is it related to the soldier?",

\quad "A1": "No"},

\quad 2: {
"Q2": "Is it related to the school?",

\quad "A2": "Yes"}
}
}
}

OUTPUT: $\langle$WORD$\rangle$ Is it something edible?\\

......

\sethlcolor{image}\hl{Example2}: ......

\sethlcolor{image}\hl{Example3}: ......
\\

\sethlcolor{prompt}\hl{Referring to the above example, please use the latest INPUT information and the pictures provided below to think divergently and ask a general question (i.e., a question with a yes or no answer) for possible $\langle$WORD$\rangle$ to help generate $\langle$WORD$\rangle$. Please note that only general questions are output.}

\sethlcolor{response}\hl{
INPUT: 

"IMAGE": "$\langle$image$\rangle$",

"IMAGE CAPTION": "$\langle$caption$\rangle$",

"RESPONSE": "$\langle$response$\rangle$",

"TIPS": $\langle$tips$\rangle$

OUTPUT:
}

\end{minipage}
\end{mdframed}

\newpage

\section*{Appendix C: Answer the question $Q_t$ and provide $A_t$}

\begin{mdframed}[backgroundcolor=gray!8]
\begin{minipage}{\linewidth}

\sethlcolor{prompt}\hl{Given a $\langle$WORD$\rangle$ and a question QUESTION about $\langle$WORD$\rangle$, you need to give the answer to this question QUESTION through common sense and reasoning. If it is correct, output Yes in the format, if it is wrong, output No in the format. Here are some examples,}\\

\sethlcolor{image}\hl{Example1}:\\

INPUT: {

"$\langle$WORD$\rangle$": "Mentor",

"QUESTION": 
"Is $\langle$WORD$\rangle$ related to soldiers?"

}

OUTPUT: \sethlcolor{prompt}\hl{No}\\

\sethlcolor{image}\hl{Example2}:\\

INPUT: {

"$\langle$WORD$\rangle$": "Cat",

"QUESTION": "Is $\langle$WORD$\rangle$ an animal?"

}
OUTPUT: \sethlcolor{prompt}\hl{Yes}\\

\sethlcolor{prompt}\hl{Please read the above examples carefully, and output OUTPUT strictly in the format given the new INPUT shown below}\\

\sethlcolor{response}\hl{INPUT: 

"$\langle$WORD$\rangle$": "$\langle$word$\rangle$",

"QUESTION": "$\langle$question$\rangle$"

OUTPUT:}
\end{minipage}
\end{mdframed}

\section*{Appendix D: The prompt details of DAESO}

\begin{mdframed}[backgroundcolor=gray!8]
\begin{minipage}{\linewidth}

\sethlcolor{prompt}\hl{Given a detailed text description of an image $\langle$IMAGE CAPTION$\rangle$, it has a very humorous and creative caption $\langle$GTR$\rangle$ and its detailed explanation $\langle$EXP$\rangle$, please help me parse the entities, relationships and causal chains of $\langle$EXP$\rangle$. 

And analyze, if $\langle$GTW$\rangle$ in $\langle$GTR$\rangle$ is replaced with $\langle$RESPONSE$\rangle$, does $\langle$RESPONSE$\rangle$ still meet the analysis, that is, does it still have similar humor, creativity and function?

......

Please answer strictly in the following format:
}\\

......

\sethlcolor{image}\hl{Example1}: ......

\sethlcolor{image}\hl{Example2}: ......
\\

{

\quad \sethlcolor{condition}\hl{"SUMMARY":} \sethlcolor{prompt}\hl{"Yes/No"},

\quad \sethlcolor{condition}\hl{"EXPLANATION":} \sethlcolor{prompt}\hl{"..."}
}

\sethlcolor{prompt}\hl{
The content of SUMMARY is Yes or No, indicating whether $\langle$RESPONSE$\rangle$ still has similar sense of humor and creativity after $\langle$GTW$\rangle$ in $\langle$GTR$\rangle$ is replaced with $\langle$RESPONSE$\rangle$;

The content of EXPLANATION is the analysis of the entities, relationships and causal chains of the paragraph $\langle$EXP$\rangle$, as well as the analysis of whether $\langle$GTR$\rangle$ still has similar sense of humor and creativity.

If $\langle$RESPONSE$\rangle$ is not a simple phrase or sentence, but a complex format, then SUMMARY is No.
}
\end{minipage}
\end{mdframed}

\end{document}